\documentclass[letterpaper, 10 pt, conference]{ieeeconf}  

\IEEEoverridecommandlockouts                              

\usepackage{times}
\usepackage{epsfig}
\usepackage{graphicx}
\usepackage{amsmath}
\usepackage{amssymb}
\usepackage{subcaption}
\usepackage{url}
\usepackage{multirow}
\usepackage{colortbl}
\usepackage[dvipsnames]{xcolor}
\usepackage{verbatim}
\usepackage{color}

\usepackage{multirow}
\usepackage{booktabs}
\usepackage{graphicx} 
\usepackage{tabularx} 
\usepackage{array,multirow,graphicx}
\usepackage{tabu}
\usepackage{booktabs} 
\usepackage{tablefootnote} 

\usepackage[breaklinks=true,bookmarks=false,colorlinks,urlcolor=black, backref=page, linktocpage=true]{hyperref}

\usepackage{caption}

\usepackage{xfrac}

\newcommand\email[1]{\href{mailto:#1}{\nolinkurl{#1}}}

\usepackage[normalem]{ulem}
\definecolor{mygray}{gray}{0.5}
\newcommand{\clinegray}[2]{\arrayrulecolor{mygray}\cline{#1-#2}\arrayrulecolor{black}}

\def\eg{\emph{e.g.}} 
\def\ie{\emph{i.e.}}

\def\etal{\emph{et al.}}

\definecolor{myRed}{RGB}{250, 50, 0}

\definecolor{myBlue}{RGB}{0, 50, 250}

\title{\LARGE \bf RomniStereo: Recurrent Omnidirectional Stereo Matching}

\author{Hualie Jiang, Rui Xu, Minglang Tan and Wenjie Jiang 
\thanks{The authors are all with Insta360 Research,  Shenzhen 518000, China, {\tt\small \{jianghualie, jerry1, tanminglang, jerett\} @insta360.com}.}
}

\begin{document}

\maketitle
\thispagestyle{empty}
\pagestyle{empty}

\begin{abstract}

Omnidirectional stereo matching (OSM) is an essential and reliable means for $360^{\circ}$ depth sensing.
However, following earlier works on conventional stereo matching, prior state-of-the-art (SOTA) methods rely on a 3D encoder-decoder block to regularize the cost volume, causing the whole system complicated and sub-optimal results. 
{Recently, the Recurrent All-pairs Field Transforms (RAFT) based approach employs the recurrent update in 2D and has efficiently improved image-matching tasks, \ie, optical flow, and stereo matching. To bridge the gap between OSM and RAFT, we mainly propose an opposite adaptive weighting scheme to seamlessly transform the outputs of spherical sweeping of OSM into the required inputs for the recurrent update, thus creating a recurrent omnidirectional stereo matching (RomniStereo) algorithm. Furthermore, we introduce two techniques, \ie, grid embedding and adaptive context feature generation, {which also contribute to RomniStereo's performance}. Our best model improves the average MAE metric by 40.7\% over the previous SOTA baseline across five datasets. When visualizing the results, our models demonstrate clear advantages on both synthetic and realistic examples. }
{The code is available at \url{https://github.com/HalleyJiang/RomniStereo}.} 

\end{abstract}


\section{INTRODUCTION}
\label{sec:intro}

{Omnidirectional depth} sensing eliminates blind spots and thus facilitates robot navigation much more significantly than conventionally limited field-of-view (FoV) pinhole perception~\cite{de2018eliminating}. Compared with the expensive acquisition means with LiDAR~\cite{liao2022kitti}, using large FoV images for estimation has more potential to produce fast and dense omnidirectional  depth~\cite{da20223d}. Thus, {omnidirectional} depth estimation has been recently investigated from various image sources, \eg, single panorama~\cite{zioulis2018omnidepth, wang2020bifuse, jiang2021unifuse, li2022omnifusion}, dual fisheyes~\cite{gao2017dual} or panoramas~\cite{wang20icra}, and quadruple fisheyes~\cite{won2019sweepnet, won2019omnimvs, won2020end}.

\begin{figure}[t]
\vspace{0pt}
\begin{center}
\includegraphics[width=\linewidth]{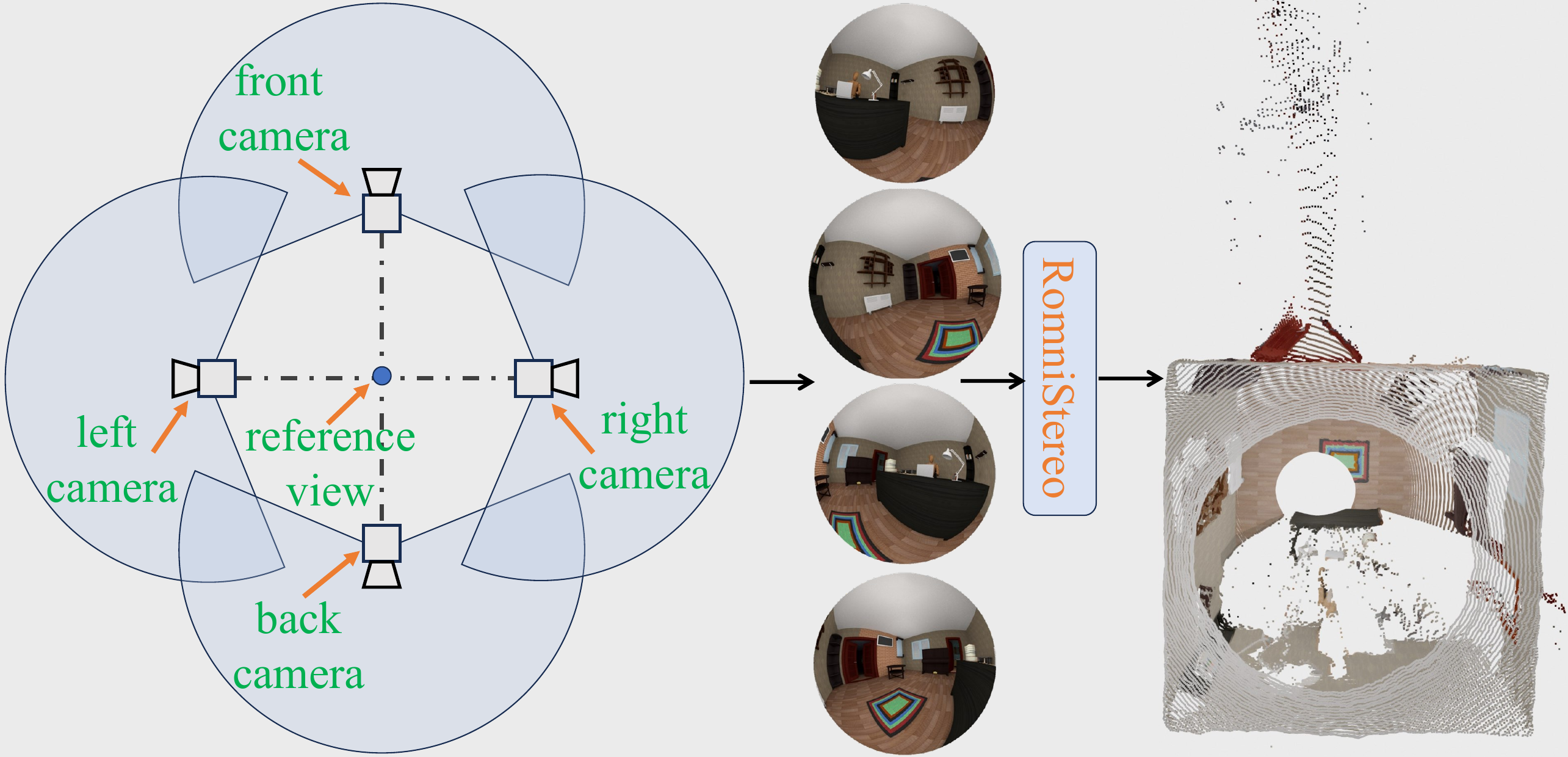}
\end{center}
\vspace{-5pt}
\caption{The illustration of the quadruple fisheye camera system and the functionality of our proposed RomniStereo. RomniStereo utilizes the four fisheye images from cameras to predict a panoramic depth map from the virtual reference view; omnidirectional reconstruction can be obtained.}
\label{fig:rig}
\vspace{-12pt}
\end{figure}

{Among different vision-based schemes, the rig of four fisheye cameras mounted on the corners of a wide square and facing outwards proposed by Won~\etal~\cite{won2019sweepnet} is feasible and reliable, as shown in Fig.~\ref{fig:rig}. The camera FoV has to be larger than $180^{\circ}$ (set defaulted to $220^{\circ}$ by Won~\etal) to ensure most points in the world are captured by at least two cameras so that matching can be performed.
The goal of OSM is to predict the omnidirectional scene reconstruction from the center of the four cameras of the rig. } Lying on the same plane makes the rig easy to install on robots. Furthermore, the multiple large FoV cameras ensure the surrounding world {can be observed by at least two cameras and the wide baseline enables the depth to be accurately triangulated. }

{Due to wide baselines, a large FoV, and frequent occlusions, accurate matching is challenging; moreover, the rig introduces multiple matches. SOTA methods, such as OmniMVS~\cite{won2019omnimvs, won2020end} and S-OmniMVS~\cite{chen2023s}, rely on an encoder-decoder block of 3D convolutions. This design is inspired by conventional stereo matching methods (\eg, GC-Net~\cite{kendall2017end} and PSMNet~\cite{chang2018pyramid}) and serves to regularize the concatenated feature volume derived from the deep feature maps of the four fisheye cameras using a spherical sweeping approach. 3D convolutions are computationally expensive and can limit efficiency. In contrast, dense image matching tasks have been revolutionized by RAFT~\cite{teed2020raft}, which primarily utilizes a 2D recurrent update module for iterative motion prediction. The advantages of RAFT have been showcased in applications such as optical flow~\cite{teed2020raft, sui2022craft} and stereo matching~\cite{lipson2021raft, li2022practical, jing2023uncertainty}.} {Therefore, the incorporation of a recurrent update module into OSM could be promising.}

\begin{figure*}[t]
\vspace{0pt}
\begin{center}
\includegraphics[width=0.99\linewidth]{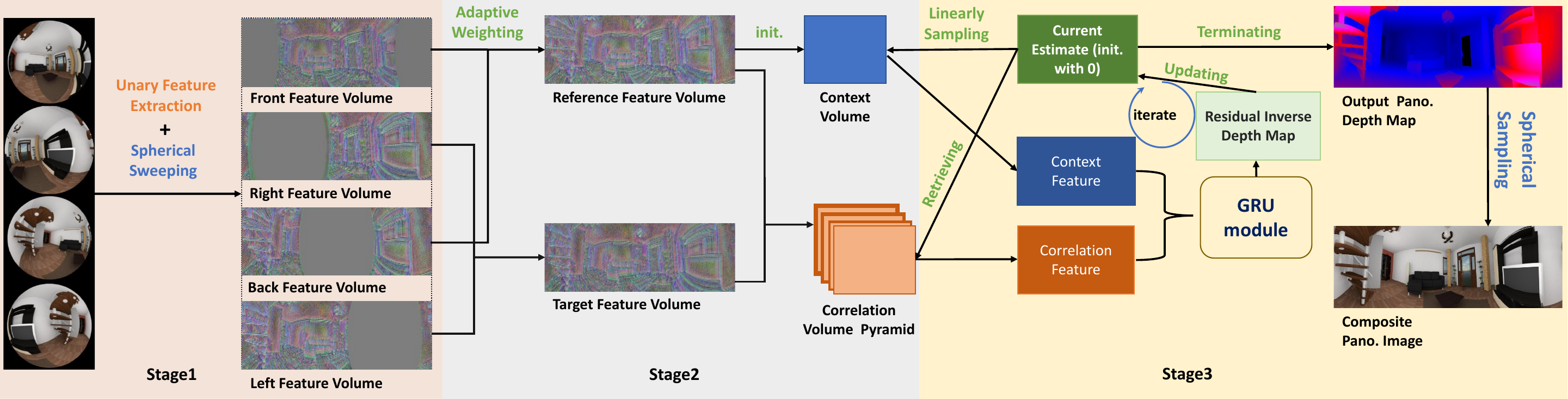}
\end{center}
\vspace{-5pt}
\caption{Our Proposed Recurrent Omnidirectional Stereo Matching Framework.}
\label{fig:framework}
\vspace{-12pt}
\end{figure*}

{However, there is a gap between the spherical sweeping feature volumes and the required inputs for the recurrent update. This recurrent update is implemented with a 2D convolutional  Gate Recurrent Unit (GRU), whose inputs are a context feature map extracted from the reference view and a correlation feature map, as illustrated in Stage3 of Fig.~\ref{fig:framework}. The context map cannot be directly obtained as there is no physically existing reference view. The correlation feature map is usually sampled from a pyramid of correlation volumes between the reference and target views. None of the four fisheye cameras are suitable to serve as the reference/target view as they cannot cover $360^{\circ}$ FoV individually. }

{We propose methods to transform raw feature volumes from spherical sweeping into the format required by the recurrent update, as shown in Stage2 of Fig.~\ref{fig:framework}. The initial step involves combining the four raw feature volumes, each with an incomplete FoV, into the reference and target volumes separately. These two volumes form the basis for computing the correlation volume using a dot product. 
For the correlation volume to be meaningful, the reference and target volumes must satisfy two criteria: they must cover the entire FoV and be distinct from one another. Since the opposite fisheye pairs cover the full FoV, we suggest adaptively weighting two opposing pairs of feature volumes to form the reference and target volumes, respectively. Given that the reference and target volumes originate from different sources, they inherently differ. Besides, we demonstrate that this opposite adaptive weighting method is superior to two other combination methods: one being tighter and the other looser.}
Motivated by the image coordinate grid in spherical sweeping containing position information, we embed it in computing the combination weights. {Additionally, rather than using a zero context feature map, we utilize an adaptive approach, \ie, initializing the context volume with the reference volume, and sampling the context feature from the context volume with the current estimate adaptively.} {We test our RomniStereo on five benchmark datasets: Omnithings, OmniHouse, Sunny, Cloudy, and Sunset. Experiments verify the effectiveness and efficiency of our method.}

Our contributions can be summarized as follows: (1) 
We introduce RomniStereo, an end-to-end recurrent omnidirectional stereo matching architecture that estimates $360^{\circ}$ depth from four orthogonal fisheye cameras. RomniStereo achieves SOTA performance, with improved accuracy and faster speed.
(2) We implement RomniStereo mainly via the proposed opposite adaptive weighting strategy, which is verified to exceed the other two alternative methods. (3) We also introduce two extra beneficial techniques to enhance RomniStereo's performance: grid embedding and adaptive context generation with reference volume initialization.

\section{RELATED WORK}
\label{sec:rw}

\subsection{Omnidirectional Depth Estimation}

The most straightforward way to obtain omnidirectional depth is single panoramic image depth estimation with purely data-driven methods. Omnidepth~\cite{zioulis2018omnidepth} is a seminal work of this aspect, which introduces a panoramic RGBD dataset, 3D60, and adopts rectangle filters to handle the distortion of equirectangular representation of panoramas. 
{Later monocular omnidirectional depth estimation has been continuously improved by new neural architectures~\cite{wang2020bifuse, jiang2021unifuse, li2022omnifusion}.}
However, the pure data-driven monocular methods require a large size of real data for training and are not geometrically reliable.

{Using stereo views to form a baseline for triangulation is more reliable. }
Gao~\etal proposed a setup of two fisheye cameras vertically mounted back-to-back~\cite{gao2017dual}. The camera FoV is $245^{\circ}$ and thus there is $65^{\circ}$ vertical overlap for depth sensing. They extracted multiple pinhole images in different horizontal angles and performed vertical stereo-matching via traditional block matching to obtain omnidirectional depth.  However, this setup requires very large FoV cameras to ensure sufficient overlap, and the mounting manner means that it can be only applied on UAVs. 
Differently, 360SD-Net~\cite{wang20icra} proposes to set two consumer $360^{\circ}$ cameras (Insta360\textsuperscript{\textregistered} ONE X) and perform vertically stereo matching via constructing a learnable cost volume from deep feature maps and using stacked hourglass 3D convolutions to regress disparity. 
However, the camera pair has to be fixed in a setup such as a pod, which makes occlusion for imaging and is difficult to install on autonomous robots. Both are affected by vertical structures in the environment. 

Won~\etal~introduced a more deploying-friendly four fisheye camera structure (Fig.~\ref{fig:rig}) for OSM~\cite{won2019sweepnet}. They first proposed SweepNet~\cite{won2019sweepnet}, which extracts feature maps from projected images from fisheye to spherical space via spherical sweeping to construct cost volume and use Semi-Global Matching (SGM)~\cite{hirschmuller2008stereo} on the cost volume to obtain the depth map. Later, they proposed to extract deep features from fisheye images and project to spherical space to construct cost volume and replace SGM with a 3D encoder and decoder block to filter the cost volume,  thus forming the end-to-end OmniMVS~\cite{won2019omnimvs}. Finally, they presented OmniMVS${}^+$~\cite{won2020end}, which adds an uncertainty estimation supervised with an entropy boundary loss. 
To help the exceptional case that cameras are not fixed well and the view directions are out of the plane, Komatsu~\etal~presented IcoSweepNet with icosahedron-based spherical sweeping~\cite{komatsu2020360}. However, the experiments show that IcoSweepNet underperforms OmniMVS until the tilt angle reaches $30^{\circ}$. Meanwhile, Meuleman~\etal~\cite{meuleman2021real}~introduced a fast traditional stereo matching algorithm to obtain $360^{\circ}$ RGBD from multiview {fisheye} images. However, this algorithm emphasizes panorama synthesis and requires the camera baseline to be short. {Recently, S-OmniMVS~\cite{chen2023s} improved OmniMVS by introducing extra modules to fisheye image feature extraction and spherical cost aggregation to make them more robust distortion. Although S-OmniMVS performs better than OmniMVS, it inevitably increases model complexity.} 
In contrast, we focus on developing an efficient and effective model for OSM.

\subsection{Pinhole Stereo Matching}

{Stereo matching from rectified pinhole images has been a traditional computer vision task for decades. 
SGM~\cite{hirschmuller2008stereo} is a representative conventional algorithm with low computational overhead and good accuracy. However, stereo matching has been significantly improved by the modern models. Mayer~\etal~\cite{mayer2016large} proposed the end-to-end DispNet that directly regresses disparity from the 2D feature maps. }Better results are obtained by GC-Net~\cite{kendall2017end}, which replaces SGM with 3D convolutions for incorporating contextual information over the cost volume from which differentiable disparity can be regressed. 
PSMNet~\cite{chang2018pyramid} further improves the 3D block by introducing spatial pyramid pooling to enlarge the receptive field and constructing a stacked hourglass 3D
encoder-decoder architecture. 
Conventionally, in the field of OSM, OmniMVS~\cite{won2019omnimvs, won2020end} uses a 3D encoder-decoder architecture to regularize the cost volume.

Recently, the image matching field has been revoluted by a new paradigm, RAFT~\cite{teed2020raft}, which is first applied to optical flow estimation and then extended to stereo matching in RAFT-Stereo~\cite{lipson2021raft} and CREStereo~\cite{li2022practical}. The idea of RAFT is to recurrently update the image motion prediction with a 2D GRU module whose inputs are the sampled correlation feature from the correlation volume over the feature maps of the reference and target views and the context map from the reference view only. As the 3D block is not required for regularization, RAFT is efficient. The effectiveness and robustness are also demonstrated in the Robust Vision Challenge 2022\footnote{\url{http://www.robustvision.net/leaderboard.php}}, where RAFT-related methods dominate the tasks of stereo matching (first three are CREStereo++\_RVC~\cite{jing2023uncertainty}, iRaftStereo\_RVC~\cite{jiang2022improved} and raft+\_RVC) and optical flow (first two are MS\_RAFT+\_RVC and RAFT-it+\_RVC). Therefore, in this paper, we take advantage of the idea of RAFT and extend this advanced paradigm to OSM.

\section{Methodology}
\label{sec:method}

Our framework, shown in Fig.~\ref{fig:framework}, consists of three stages. The first stage includes unary feature extraction and spherical sweeping, which are borrowed from OmniMVS~\cite{won2020end}. The second stage, contributed by us, involves adaptive feature volume generation. This stage acts as a bridge between the outputs of spherical sweeping and the required inputs for the recurrent update. Finally, the third stage is the recurrent update, which is adapted from RAFT-Stereo~\cite{lipson2021raft}.

\subsection{Unary Feature Extraction and Spherical Sweeping}
\label{sec:pre}

{This section briefly introduces the unary feature extraction from four fisheye images and further discusses how spherical sweeping projects the raw fisheye feature map into the spherical feature volumes. }

{\textbf{Unary Feature Extraction} is a common step in image analysis with deep learning and our system uses a shared 2D convolutional neural network (CNN) to extract deep feature maps from the four fisheye images.} The capacity of the CNN is controlled by a base channel number $C$. To fairly compare with OmniMVS, we employ the same CNN structure. We denote the extracted feature maps from front, right, back, and left images as $U_f$, $U_r$, $U_b$, and $U_l$.

\textbf{Spherical sweeping} is to project the fisheye feature maps to a series of spheres centered at the reference point with $N$ predefined radii.
The sphere is evenly sampled with the spherical coordinate $\langle\theta, \phi\rangle$. The unit ray for $\langle\theta,\phi\rangle$ is $\bar{\mathbf{p}}(\theta,\phi) = (\cos(\phi)\cos(\theta),\sin(\phi),\cos(\phi)\sin(\theta))^{\top}$.
The inverse radius $d_n$ is uniformly swept from 0 to $d_{max}$, where $1/d_{max}$ is the predefined minimum depth. 
The sampled 3D point of the sphere is thus represented as $\bar{\mathbf{p}}(\theta,\phi)/d_n$. 

With the calibrated extrinsic and intrinsic parameters of the $i$-th camera, one can define a projection function, $\Pi_i$, which maps the 3D point to the 2D image coordinate. Thus, the projected image coordinate grid to $i$-th camera is, 

\begin{equation}
    G_{i}(\theta, \phi, n)=\Pi_i(\bar{\mathbf{p}}(\theta,\phi)/d_n).
    \label{eq:sph_grid}
\end{equation}

With the image coordinate grid $G_{i}$, we can perform bilinear interpolation to obtain the spherical feature map at $n$-th sphere for $i$-th image, 

\begin{equation}
 \mathcal{S}_i(\phi, \theta, n) = U_{i}\langle G_{i}(\theta, \phi, n) \rangle.
\label{eq:sph_feat}
\end{equation}

To save the memory and computation overhead, following OnmiMVS, we use every other of predefined inverse depth values, \ie, $[d_0, d_2,\ldots, d_{N-1}]$ in feature map warping. The target resolution of output depth maps is $H \times W$. Unlike OnmiMVS, we do not sample the spherical coordinate with the target output resolution. Instead, we just use a half resolution, as we can use convex upsampling~\cite{teed2020raft} to promote the prediction to the full resolution effectively. Therefore, the size of grid volume $G_{i}$ and spherical feature volume $\mathcal{S}_i$ are $H/2 \times W/2 \times N/2 \times 2$ and $H/2 \times W/2 \times N/2 \times C$. Note that the fisheye image cannot cover the sphere space. For the 3D point outside of the FOV of the fisheye image, the corresponding values in $\mathcal{S}_i$ are set to 0, {which corresponds to the homogeneous grey region in the resulting spherical feature volumes of spherical sweeping (Fig.~\ref{fig:framework}). }

\subsection{Adaptive Feature Volume Generation}
\label{sec:afvg}

{The recurrent update in RAFT requires a deep correlation volume between the reference and target views to provide a correlation map sampled with the current estimate. Meanwhile, the recurrent update also requires an extra context feature map extracted from the reference view as input. Two demanded inputs for the recurrent update are inconsistent with the output spherical feature volumes from spherical sweeping. To bridge the inconsistency, we first propose to separately combine the four spherical feature volumes with incomplete FoV into the reference and target volumes, from which the correlation volume can be computed via dot product. 
As discussed in Sec.~\ref{sec:intro}, the reference and target volumes must cover the whole FoV and be different. }

\begin{table}[t]
\vspace{0pt}
  \centering
  \resizebox{0.5\textwidth}{!}{
  \begin{tabular}{ l |cccc}
  \toprule
  {Methods} & input nodes & hidden nodes & output nodes & output activation \\
  \hline
  Opp. Ada. Weighting  & $2C$ & $C$ & $1$ & sigmoid \\
  + Grid Embedding  & $2C+4$ & $C$ & $1$ & sigmoid \\
  \clinegray{1}{5}
  All-Weighting  & $4C$ & $C$ & $4$ & softmax \\
  \bottomrule
  \end{tabular}
  }
  \vspace{-2pt}
\caption{MLPs of different volume generation methods.}
\vspace{-12pt}
\label{tab:mlp}
\end{table}

{To generate a reference volume or a target volume that covers the complete space efficiently, we argue that one has to use the minimal number of raw spherical feature volumes to save computation overhead. Using only one raw spherical feature volume is impossible, as none of the four raw spherical feature volumes cover the complete space. From the camera rig (Fig.~\ref{fig:rig}), we find that the front and back cameras together cover a $360^{\circ}$ FoV, which is also illustrated by the front spherical volume and back spherical volume in Fig.~\ref{fig:framework}. This is also true for the left and right cameras. Therefore, it is possible to use two spherical volumes to generate either the reference volume or the target volume. We thus proposed an opposite combination strategy, \ie, using the front and back volumes to generate the reference volume and the left and right volumes to generate the target volume. In this way, the reference volume and the target volume come from different sources, they are different by nature.  }

{A feasible way to implement the above strategy is to compute a weighted sum of the opposite volumes as the reference/target volume. To ensure numerical stability, the summation of the two weight volumes should be 1 at every entry.  To achieve this target and simultaneously make the weight volume adaptive to the opposite volumes, we use the concatenation of two opposite volumes to predict a weight volume for one side, and the 1 minus the predicted volume can be the weight volume for the other side. We denote our method as \textbf{opposite adaptive weighting}. }

{Technically, we use a {Multi-Layered Perceptron (MLP)} with \textit{sigmoid} activation to compute the weight volume. As the channels of concatenated opposite volumes are $2C$, the input layer's nodes of the MLP are $2C$, and the output layer's node is set to 1 as only one weight to predict. We only use one hidden layer, whose nodes are proportionally set to $C$. Furthermore, considering that the image coordinate grid $G_{i}$ (Eqn.~\ref{eq:sph_grid}) in spherical warping represents the positional information in the fisheye image, we further incorporate the informative $G_{i}$ with $\mathcal{S}_i$ to the MLP to generate better weight volumes (denoted as \textbf{grid embedding}). As the valid region in $G_{i}$ is normalized to $[-1, 1]$ in bilinear interpolation, we set the invalid region to $-2$ to ensure numerical stability. Grid embedding slightly increases the input layer's nodes to $2C+4$, as listed in Tab~\ref{tab:mlp}. }

For the front volume $\mathcal{S}_f$ and the back volume $\mathcal{S}_b$, denoting the weight volume of $\mathcal{S}_f$ as $\mathcal{W}_f$ ($H/2 \times W/2 \times N/2 \times 1$), then the reference volume is computed as,
$\mathcal{S}_b$, 
\begin{equation}
 \mathcal{S}_{ref} = \mathcal{W}_f\odot\mathcal{S}_f+(1-\mathcal{W}_f)\odot\mathcal{S}_b.
\label{eq:sph_ref_feat}
\end{equation}
Similarly, we use another MLP to generate the weight volume $\mathcal{W}_r$ for $\mathcal{S}_r$ and analogously compute the target volume.

\begin{figure}[!ht]
\vspace{0pt}
\centering
\resizebox{\linewidth}{!}{
\newcommand{\turnheightnew}{0.176\columnwidth}

\centering

\renewcommand{\arraystretch}{0.5}
\begin{tabular}{@{\hskip 1mm} c@{\hskip 1mm}c@{\hskip 1mm}c@{\hskip 1mm}c@{}}

{\rotatebox{90}{\hspace{3mm} \Large $d_0$}} &
\includegraphics[height=\turnheightnew]{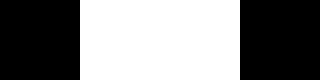} &
\includegraphics[height=\turnheightnew]{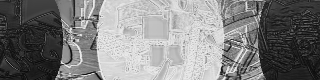} &
\includegraphics[height=\turnheightnew]{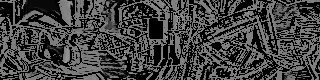} \\

{\rotatebox{90}{\hspace{1.2mm} \Large $d_{N/2}$}} &
\includegraphics[height=\turnheightnew]{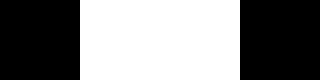} &
\includegraphics[height=\turnheightnew]{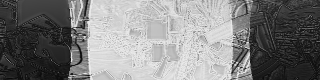} &
\includegraphics[height=\turnheightnew]{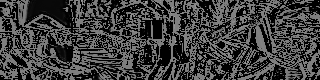} \\

{\rotatebox{90}{\hspace{1mm} \Large $d_{N-1}$}} &
\includegraphics[height=\turnheightnew]{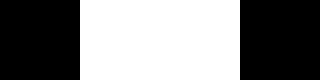} &
\includegraphics[height=\turnheightnew]{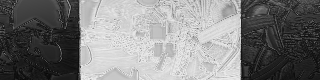} &
\includegraphics[height=\turnheightnew]{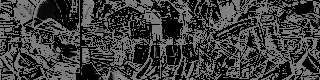} \\

&
\Large Opposite Interleving &
\Large Opposite Adaptive Weighting&
\Large All-Weighting \\

\end{tabular}
 }
\vspace{-1pt}
\caption{The weighting masks for $\mathcal{S}_f$ of different methods.}
\label{fig:masks}
\vspace{-12pt}
\end{figure}

{Alternative methods to the opposite adaptive weighting method exist.} A naive approach to implement the opposite combination is the \textbf{opposite interleaving}~\cite{won2020end}, \ie, interleaving the middle half of $\mathcal{S}_f$ with the bilateral quarter of $\mathcal{S}_b$ as the reference volume and the left half of $\mathcal{S}_l$ and the right half of $\mathcal{S}_r$ as the target volume. However, this method inevitably causes discontinuity in the boundary of interleaving. {The equivalent weight mask of opposite interleaving is binary, shown in the first column of Fig.~\ref{fig:masks}, which is fixed for different $d_n$, while the masks of the adaptive weighting method in the second column can adaptively change with $d_n$ and scene structures. The adaptive approach combines the opposite volumes more smoothly than the strict interleaving one, thus producing better reference/target volumes. }

There is a more loose combination method, \ie, generating reference/target volumes by weighting all the spherical feature volumes (denoted as \textbf{all-weighting}). To implement this method, we input the concatenation of all of the four spherical feature volumes to two MLPs with \textit{softmax} activation to output the 4-channel reference and target weight volumes, which are then used to combine all $\mathcal{S}_i$ to generate the reference and target volumes. {The information of the MLP of all-weighting is also listed in Tab~\ref{tab:mlp}.} However, this design does not consider the inductive bias from the camera structure, \ie, the opposite cameras can cover the full $360^{\circ}$ field, and we find the learned weighting maps also do not reflect the structure (third column of Fig.~\ref{fig:masks}).

{Despite differences in reference and target volumes computation methods, using them to construct the correlation is the same, \ie, computing their inner product,}
\begin{equation}
 \mathcal{C} = \mathcal{S}_{ref} \cdot \mathcal{S}_{tgt}, 
\label{eq:corr}
\end{equation}
whose size is $H/2 \times W/2 \times N/2$. 

{As the recurrent update of RAFT-Stereo requires the sampled correlation feature map to have a large receptive range in the disparity direction, a 4-level pyramid of correlation volumes has to be built. The other 3 lower scale correlation volumes can be obtained by repeatedly downsampling $\mathcal{C}$ by half in the last dimension 3 times, and the size of \textit{i}-th correlation volume $\mathcal{C}_i$ is $H/2 \times W/2 \times N/2^i$. }

{With the availability of a FoV-complete reference volume, the generation of the context feature map is also easy to implement. We propose to initialize a context feature volume with the reference volume and linearly sample the context feature from the context volume with the current estimate. Such a context map adaptively changes with iterations. Thus, we call it an \textbf{adaptive context} map. Ablation experiments show that the dedicated adaptive context map is better than a naive fixed zero context map. }

\subsection{Recurrent Update}
\label{sec:reu}

{The recurrent update is about Stage3 in Fig.~\ref{fig:framework}, where the main neural module is a 2D convolutional GRU. To match the complexity of the feature extractor in Stage1 and the MLPs in Stage2, the hidden dimension of GRU is set to $2C$. The hidden state of GRU is initialized by projecting the linearly sampled context map from the context volume with the zero inverse depth map from a $C$-channel one to a $2C$-channel one with a $1\times{1}$ 2D convolution layer. }

{Let's set that the number of recurrent iterations is $M$. The iteration starts from the current inverse depth estimate, $\mathbf{d}^i, i=\{0, 1, \cdots, M\}$, where $\mathbf{d}^0$ is initialized with 0.} {Given the current estimate, the next step is looking up from the context volume and the correlation volume pyramid provided to produce the inputs required by the GRU module, \ie, the context feature map and the correlation feature map. Similar to the generation of the initial hidden state, the context map is linearly sampled from the context volume in the depth dimension with the current estimate. 
To incorporate information about the depth of the surrounding area into the correlation feature map, we not only sample the correlation corresponding to the current inverse depth from the correlation volume but also sample those of the $2r(r=4)$ nearby inverse depth values.
Furthermore, as there are 4-level pyramid correlation volumes to be retrieved, the dimension of the sampled correlation feature map is $4\times(2r+1)$. }

{Next, the GRU module inputs the sampled correlation feature map and context feature map to update its hidden state which is then used to predict a residual inverse depth map $\Delta\mathbf{d}_i$. The next estimated inverse depth is updated as}
\begin{equation}
    \mathbf{d}^{i+1} = \mathbf{d}^i+\Delta\mathbf{d}^i.
\end{equation}
{As the recurrent update is performed at $1/2$ resolution, a mask for convex upsampling~\cite{teed2020raft} is also computed from the hidden state to recover the full resolution. }

\begin{table}[t]
\vspace{0pt}
  \centering
  \resizebox{0.48\textwidth}{!}{
  \begin{tabular}{ l |ccccc}
  \toprule
  {Dataset} & OmniThings & OmniHouse & Sunny & Cloudy & Sunset \\
  \hline
  \#train  & 9216 & 2048 & 700 & 700 & 700\\
  \#test & 1024 & 512 & 300 & 300 & 300 \\
  \bottomrule
  \end{tabular}
  }
\caption{The Statistics of the Datasets.}
\vspace{-10pt}
\label{tab:datasets}
\end{table}

\subsection{Loss Function}
\label{sec:loss}

{In this section, we define the loss function to train our whole RomniStereo framework end to end. Given ground truth inverse depth index map $\mathbf{d}_{gt}$, we follow~\cite{teed2020raft, lipson2021raft} to supervise the N predictions in Sec.~\ref{sec:reu} with exponentially increasing weights. The computation of the loss is, }
\begin{equation}
    \mathcal{L} = \sum_{i=1}^{M} \gamma^{M-i} ||\mathbf{d}^{gt} - \mathbf{d}^i||_1, \qquad \text{where } \gamma=0.9 .
\end{equation}

\section{EXPERIMENTS}

\subsection{Experimental Settings}

\subsubsection{Datasets and Evaluation}

We perform experiments on the commonly used virtual datasets, \ie, OmniThings OT), OminiHouse(OH), Sunny(Sn), Cloudy(Cd), and Sunset(Ss) introduced in previous works~\cite{won2019sweepnet, won2019omnimvs, won2020end}. The OmniThings is a dataset above generic objects in various photometric and geometric environments, similar to Flythings3D~\cite{mayer2016large} for conventional stereo matching. OmniHouse is a dataset of various indoor environments. The remaining three are about driving scenarios, and they share the same scenes but have different weather conditions. The statistics of the datasets are listed in Tab.~\ref{tab:datasets}. We follow OmniMVS~\cite{won2019omnimvs, won2020end} to pre-train our models on OmniThings for 30 epochs and then fine-tune the models on OmniHouse and Sunny for 15 epochs (models with fine-tuning are marked with an ending -ft). Both pre-trained and fine-tuned models are evaluated on the test sets of all datasets. 
We follow OmniMVS to use the percent error of the predicted inverse index from GT compared to the number of predefined depth indices ($N$). The specific metrics are the percentage of errors larger than 1, 3, 5 (\textgreater{}1, \textgreater{}3, \textgreater{}5), mean absolute error (MAE), and root mean square error (RMS).
We also adopt some real-world data samples provided by~\cite{won2020end} for visual comparison.

\subsubsection{Implementation and Training Details}
To fairly compare with OmniMVS$^+$~\cite{won2020end} on both effectiveness and efficiency, we maintain {most of the} experiment settings. The number of specified sweeping spheres is set to $192$ and the predicted and GT depth maps are cropped to $H=160 (-\pi/4 \leq \pi/4)$ and $H=640 (-\pi \leq \pi)$. 
As mentioned in Sec.~\ref{sec:method}, we use the same unary feature extractor with OmniMVS and make the complexity of the latter modules proportional to that of the extractor. Therefore, we also train the proposed networks with different numbers of channels $C$, and the resulting model is named RomniStereo${}_C$.  We do not only experiment with $C=4$, $8$, and $32$ but also increase $C$ to $64$ as our model is small enough to support it. The number of recurrent iterations is set to 12 in both training and evaluation, as the number is sufficient for convergence. {Our model is implemented in Pytorch. All the model parameters are randomly initialized and trained from scratch using one RTX 3090 GPU}. 
We use the AdamW~\cite{loshchilov2018decoupled} optimizer to train our models and use one one-cycle learning rate scheduler with a maximum learning rate of $5\times 10^{-4}$.

\begin{table}
\vspace{0pt}
\centering
\resizebox{\linewidth}{!}
{
\footnotesize
\begin{tabular}{l|c c|c c|c c}
\bottomrule
\multicolumn{1}{l|}{Dataset} & \multicolumn{2}{c|}{Omnithings} &     \multicolumn{2}{c|}{Omnihouse} & \multicolumn{2}{c}{Sunny} \\
\clinegray{2}{7}
\multicolumn{1}{l|}{Metric} & \multicolumn{1}{c}{\textgreater{}1} & \multicolumn{1}{c|}{MAE} & \multicolumn{1}{c}{\textgreater{}1} & \multicolumn{1}{c|}{MAE} & \multicolumn{1}{c}{\textgreater{}1} & \multicolumn{1}{c}{MAE} \\
\hline
opposite interleaving & 38.84 & 2.81 & 26.69 & 1.60 & 20.37 & 1.22 \\
all-weighting & 41.41 & 2.94 & 25.38 & 1.48 & 20.84 & 1.24 \\
w/o grid embedding & 36.75 & 2.64 & 22.76 & 1.34 & 18.79 & 1.16 \\
w/o adaptive context & 37.46 & 2.69 & 22.80 & 1.35 & 18.11 & 1.12 \\ 
full model  & 35.61 & 2.52  & 21.82 & 1.33 & 17.34 & 1.06 \\
\toprule
\end{tabular}
}
\caption{Ablation Study.}
\vspace{-12pt}
\label{tab:ablation}
\end{table}

\subsection{Experimental Results}

\begin{table*}[tb!]
\vspace{0pt}
\centering
\resizebox{0.9\textwidth}{!}{%
\begin{tabular}{cl|rrrrrr|rrrrrc|c}
\bottomrule
\multicolumn{2}{l|}{~~Dataset} & \multicolumn{6}{c|}{OmniThings} & \multicolumn{5}{c}{OmniHouse} & &Run Time\\
\cline{3-14}
\multicolumn{2}{l|}{~~Metric} & \multicolumn{1}{c}{~\textgreater{}1~} & \multicolumn{1}{c}{~\textgreater{}3~} & \multicolumn{1}{c}{~\textgreater{}5~} & \multicolumn{1}{c}{~MAE~} & \multicolumn{1}{c}{~RMS~} & & \multicolumn{1}{c}{~\textgreater{}1~} & \multicolumn{1}{c}{~\textgreater{}3~} & \multicolumn{1}{c}{~\textgreater{}5~} & \multicolumn{1}{c}{MAE} & \multicolumn{1}{c}{~RMS~} & & (s)\\ 
\toprule 
\multicolumn{15}{l}{\it{~~Non-learning based method}} \\ 
\clinegray{1}{15}
\multicolumn{2}{l|}{~~Sphere-Stereo~\cite{meuleman2021real}} & 80.01 & 56.67 & 44.06 & 9.14 & 14.06 & & 65.84 & 27.29 & 12.84 & 2.82 & 4.60 & & 0.21 \\
\toprule 
\multicolumn{15}{l}{\it{~~Trained on OmniThings only}} \\
\clinegray{1}{15}
\multicolumn{2}{l|}{~~OmniMVS${}^+_4$~\cite{won2020end}}      & 46.01 & 21.00 & 13.59 & 2.97 & 6.48 & & 37.77 & 13.80 & 7.43 & 1.88 & 3.93 & & 0.11 \\
\multicolumn{2}{l|}{~~\textbf{RomniStereo${}_4$}}             & 35.61 & 17.05 & 11.46 & 2.52 & 6.13 & & 21.82 & 9.24  & 5.67  & 1.33 & 2.96 & & 0.09 \\
\multicolumn{2}{l|}{~~OmniMVS${}^+_8$~\cite{won2020end}}      & 32.26 & 13.36 & 8.67 & 2.05 & 5.21 & & 29.52 & 10.34 & 5.96 & 1.62 & 3.53 & & 0.19 \\
\multicolumn{2}{l|}{~~\textbf{RomniStereo${}_8$}}             & 28.67 & 12.90 & 8.64 & 1.99 & 5.31 & & 20.02 & 8.00 & 4.70 & 1.17 & 2.66 & & 0.10 \\
\multicolumn{2}{l|}{~~OmniMVS~\cite{won2019omnimvs}}          & 47.72 & 15.12 &  8.91 & 2.40 & 5.27 & & 30.53 & 10.29 & 6.27 & 1.72 & 4.05 & & 0.82 \\
\multicolumn{2}{l|}{~~S-OmniMVS~\cite{chen2023s}}              & 28.03 & 10.40 & 6.33 & 1.48 & \bf 3.68 & & 18.86 & 8.05 & 4.90 & 1.06 & 2.41 & & - \\
\multicolumn{2}{l|}{~~OmniMVS${}^+_{32}$-IS~\cite{won2020end}}  & 24.11 &  9.38 &  5.84 & 1.45 & 4.14 & & 23.91 &  8.97 & 5.63 & 1.41 & 3.33 & & 0.72 \\
\multicolumn{2}{l|}{~~OmniMVS${}^+_{32}$~\cite{won2020end}}     & 20.70 &  \underline{8.18} &  \underline{5.49} & \underline{1.37} & 4.11 & & 19.89 &  5.89 & 3.99 & 1.30 & 2.64 & & 0.82 \\
\multicolumn{2}{l|}{~~\textbf{RomniStereo${}_{32}$}}              & \underline{20.42} & 8.49 & 5.81 & 1.39 & 4.22 & & \underline{12.13} & \underline{4.73} & \underline{3.02} & \underline{0.80} & \underline{1.85} & & 0.21 \\
\multicolumn{2}{l|}{~~\textbf{RomniStereo${}_{64}$}}              &\bf 17.77 &\bf 7.52 &\bf 5.00 &\bf 1.22 & \underline{3.90} & & \bf10.52 & \bf4.05 & \bf2.69 & \bf0.74 & \bf1.73 & & 0.44 \\

\toprule 
\multicolumn{15}{l}{\it{~~Finetuned on OmniHouse and Sunny}} \\
\clinegray{1}{15}
\multicolumn{2}{l|}{~~OmniMVS${}^+_4$-ft~\cite{won2020end}}   & 53.99 & 35.38 & 27.57 & 5.68 & 9.98 & & 15.40 & 5.00  & 2.85 & 0.86 & 1.98 & & 0.11 \\
\multicolumn{2}{l|}{~~\textbf{RomniStereo${}_4$-ft}}          & 50.01 & 33.22 & 26.30 & 5.38 & 9.59 & & 11.45 & 4.52 & 2.89 & 0.77 & 1.92 & & 0.09 \\
\multicolumn{2}{l|}{~~\textbf{RomniStereo${}_8$-ft}}          & 44.50 & 28.61 & 22.05 & 4.43 & 8.46 & & 8.66 & 3.36 & 2.14 & 0.59 & 1.56 & & 0.10 \\
\multicolumn{2}{l|}{~~OmniMVS-ft~\cite{won2019omnimvs}}       & 50.28 & 22.78 & 15.60 & 3.52 & 7.44 & & 21.09 &  4.63 & 2.58 & 1.04 & 1.97 & & 0.82 \\ 
\multicolumn{2}{l|}{~~S-OmniMVS-ft~\cite{chen2023s}}   & - & - & - & - & - & & 6.99 & \bf 1.79  & \bf 0.97 & \bf 0.42 & \bf 1.06 & & - \\

\multicolumn{2}{l|}{~~OmniMVS${}^+_{32}$-ft~\cite{won2020end}}  & 44.79 & 27.17 & 20.41 & 4.23 & 8.42 & &  9.70 &  3.51 & 2.13 & 0.64 & 1.69 & & 0.82 \\
\multicolumn{2}{l|}{~~\textbf{RomniStereo${}_{32}$-ft}}           & \underline{34.32} & \underline{19.76} & \underline{14.22} & \underline{2.81} & \underline{6.47} & & \underline{6.02} & {2.49} & {1.73} &{0.49} &{1.31} & & 0.21 \\
\multicolumn{2}{l|}{~~\textbf{RomniStereo${}_{64}$-ft}}           & \bf 29.84 & \bf{16.21} & \bf 11.28 & \bf 2.26 & \bf 5.60 & &\bf 5.28 & \underline{2.22} & \underline{1.51} &\bf 0.42 & \underline{1.14} & & 0.44 \\

\toprule
\end{tabular}
}

\vspace{3pt}
\resizebox{\textwidth}{!}{%
\begin{tabular}{l|rrrrr|rrrrr|rrrrr} \bottomrule
\multicolumn{1}{l|}{Dataset} & \multicolumn{5}{c|}{Sunny} & \multicolumn{5}{c|}{Cloudy} & \multicolumn{5}{c}{Sunset} \\
 \cline{2-16}
\multicolumn{1}{l|}{Metric} & \multicolumn{1}{c}{\textgreater{}1} & \multicolumn{1}{c}{\textgreater{}3} & \multicolumn{1}{c}{\textgreater{}5} & \multicolumn{1}{c}{MAE} & \multicolumn{1}{c|}{RMS} & \multicolumn{1}{c}{\textgreater{}1} & \multicolumn{1}{c}{\textgreater{}3} & \multicolumn{1}{c}{\textgreater{}5} & \multicolumn{1}{c}{MAE} & \multicolumn{1}{c|}{RMS} & \multicolumn{1}{c}{\textgreater{}1} & \multicolumn{1}{c}{\textgreater{}3} & \multicolumn{1}{c}{\textgreater{}5} & \multicolumn{1}{c}{MAE} & \multicolumn{1}{c}{RMS} \\ 

\toprule \multicolumn{16}{l}{\it{Non-learning based method}} \\ 
\clinegray{1}{16}
Sphere-Stereo~\cite{meuleman2021real} & 76.46 & 45.99 & 28.46 & 4.92 & 8.35 & 77.57 & 47.08 & 28.39 & 4.50 & 7.21 & 77.38 & 46.11 & 28.49 & 5.15 & 8.89 \\

\toprule 
\multicolumn{16}{l}{\it{Trained on OmniThings only}} \\ 
\clinegray{1}{16}
OmniMVS${}^+_4$~\cite{won2020end}      & 26.18 & 7.06 & 4.37 & 1.24 & 3.06 & 28.50 & 6.62 & 3.93 & 1.23 & 2.92 & 25.29 & 6.92 & 4.18 & 1.22 & 3.06 \\
\textbf{RomniStereo${}_4$}   & 17.34 & 6.92  & 4.54  & 1.06 & 3.30 & 16.65 & 6.30  & 4.09  & 1.01 & 3.04 & 16.77 & 6.63  & 4.28  & 1.04 & 3.27 \\
OmniMVS${}^+_8$~\cite{won2020end}      & 18.49 & 6.13 & 3.93 & 1.10 & 3.07 & 18.85 & 5.89 & 3.72 & 1.08 & 2.94 & 17.99 & 6.08 & 3.85 & 1.09 & 3.02 \\
\textbf{RomniStereo${}_8$}   & 15.46 & 6.54 & 4.41 & 0.99 & 3.12 & 15.14 & 6.09 & 4.10 & 0.95 & 2.97 & 15.25 & 6.42 & 4.24 & 0.98 & 3.12 \\
OmniMVS~\cite{won2019omnimvs} & 27.16 & 6.13 & 3.98 & 1.24 & 3.09 & 28.13 & 5.37 & 3.54 & 1.17 & 2.83 & 26.70 & 6.19 & 4.02 & 1.24 & 3.06 \\
S-OmniMVS~\cite{chen2023s} & 17.19 & 6.03 & 3.89 & 1.11 & 3.60 & - & - & - & - & - & - & - & - & - & - \\
OmniMVS${}^+_{32}$-IS~\cite{won2020end} & 17.46 & 5.73 & 3.60 & 0.99 & 2.76 & 17.67 & 5.84 & 3.82 & 1.04 & 3.00 & 17.28 & 5.63 & 3.42 & 0.98 & 2.71 \\
OmniMVS${}^+_{32}$~\cite{won2020end}     & 13.57 & \bf 4.81 & \bf 3.10 & 0.88 & \bf 2.56 & 13.59 & \bf 4.81 & \bf 3.15 & 0.87 & 2.53 & 13.36 & \bf 4.71 & \bf 2.93 & 0.87 & \bf 2.50 \\
\textbf{RomniStereo${}_{32}$}    & \underline{12.28} & 5.59 & 3.79 & \underline{0.80} & 2.68 & \underline{11.86} & 5.08 & \underline{3.44} & \underline{0.75} & \underline{2.50} & \underline{12.30} & 5.45 & 3.48 & 0.78 & 2.67 \\
\textbf{RomniStereo${}_{64}$}  & \bf 11.25 & \underline{5.30} & \underline{3.59} & \bf 0.75 & \underline{2.57} & \bf 10.97 & \underline{5.03} & \underline{3.44} & \bf 0.73 & \bf 2.47 & \bf 10.94 & \underline{4.99} & \underline{3.29} & \bf 0.72 & \underline{2.56} \\

\toprule 
\multicolumn{16}{l}{\it{Finetuned on OmniHouse and Sunny}} \\ 
\clinegray{1}{16}
OmniMVS${}^+_4$-ft~\cite{won2020end} & 10.54 & 3.42 & 2.11 & 0.65 & 2.06 & 10.22 & 3.19 & 1.92 & 0.61 & 1.94 & 10.81 & 3.64 & 2.21 & 0.66 & 2.11 \\
\textbf{RomniStereo${}_4$-ft}  & 9.30 & 3.47 & 2.21 & 0.60 & 2.25 & 9.54 & 3.47 & 2.17 & 0.60 & 2.20 & 9.48 & 3.57 & 2.27 & 0.60 & 2.25  \\
\textbf{RomniStereo${}_8$-ft}  & 7.38 & 2.75 & 1.72 & 0.48 & 1.92  & 7.53 & 2.69 & 1.66 & 0.48 & 1.87 & 7.65 & 2.94 & 1.86 & 0.50 & 2.01 \\
OmniMVS-ft~\cite{won2019omnimvs} & 13.93 & 2.87 & 1.71 & 0.79 & 2.12 & 12.20 & 2.48 & 1.46 & 0.72 & 1.85 & 14.14 & 2.88 & 1.71 & 0.79 & 2.04 \\
S-OmniMVS-ft~\cite{chen2023s} & 6.66 & 2.18 & 1.40 & 0.47 & 1.98 & - & - & - & - & - & - & - & - & - & - \\
OmniMVS${}^+_{32}$-ft~\cite{won2020end} &  7.48 & 3.57 & 2.42 & 0.57 & 2.42 &  7.29 & 3.38 & 2.30 & 0.54 & 2.31 &  7.82 & 3.60 & 2.42 & 0.58 & 2.36 \\
\textbf{RomniStereo${}_{32}$-ft} & \underline{5.19} & \underline{1.98} & \underline{1.23} & \underline{0.36} & \underline{1.55} & \underline{5.63} & \underline{2.03} & \underline{1.29} & \underline{0.39} & \underline{1.72} & \underline{5.53} & \underline{2.13} & \underline{1.34} & \underline{0.37} & \underline{1.61} \\
\textbf{RomniStereo${}_{64}$-ft}  &\bf 4.61 &\bf 1.78 &\bf 1.10 &\bf 0.32 &\bf 1.43 &\bf 4.94 &\bf 1.83 &\bf 1.16 &\bf 0.34 &\bf 1.53 &\bf 4.88 &\bf 1.90 &\bf 1.19 &\bf 0.34 &\bf 1.49 \\

\toprule
\end{tabular}
}

\vspace{-7pt}
\caption{Quantitative Comparison. The top table shows the results of OmniThings and OmniHouse, and the running time with a Nvidia 1080Ti. The bottom table shows the results of Sunny, Cloudy, and Sunset. The best results are marked in \textbf{bold} and the second best results are marked with \underline{underline}. }
\label{tab:quan}
\vspace{-12pt}
\end{table*}

\subsubsection{Ablation Study}
We perform our ablation study on the simplest setting, \ie, $C=4$, and train the models on OmniThings for 30 epochs. The quantitative results are listed in Tab.~\ref{tab:ablation}. The top third row compares the different weighting strategies in Sec.~\ref{sec:afvg} in generating reference/target feature volumes. The first row is the results of the tightest opposite interleaving method, the second row is the loosest all weighting scheme, and the third row is our elaborate opposite adaptive weighting approach (but the grid embedding is not included for fair comparison). On the three datasets, the adaptive approach outperforms the other two variants by over $2.5\%$ on \textgreater{}1 and around $9\%$ on MAE. We also observe that in most cases, the all-weighting scheme is inferior to the simplest opposite interleaving method, indicating the importance of utilizing prior knowledge of the camera structure.

We examine the effectiveness of image coordinate grid embedding and the adaptive context in the bottom third rows. The fourth row represents using a fixed zero context map for recurrent updating instead of the proposed adaptive context feature map generation in Sec.~\ref{sec:afvg}. The last row is our full model. When ablating either of the proposed techniques, the performance drops, demonstrating their effectiveness in recurrent omnidirectional stereo matching.

\begin{figure*}[!ht]
\vspace{0pt}
\centering
\resizebox{0.99\textwidth}{!}{
\newcommand{\turnheightnew}{0.176\columnwidth}

\centering

\renewcommand{\arraystretch}{0.5}
\begin{tabular}{@{\hskip 1mm} c@{\hskip 1mm} c@{\hskip 1mm}c@{\hskip 1mm}c@{\hskip 1mm}c@{\hskip 1mm}c@{}}

{\rotatebox{90}{\hspace{1.0mm}\scriptsize OmniThings}} &
\includegraphics[height=\turnheightnew]{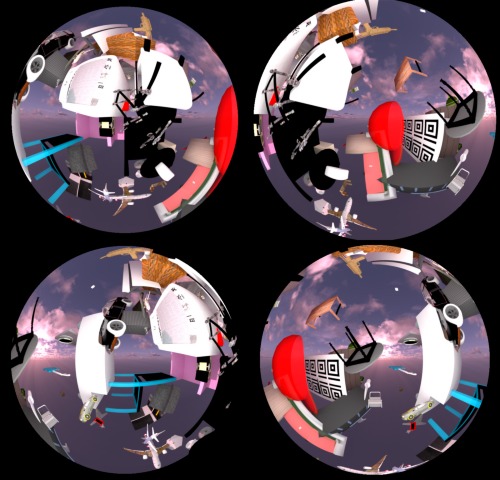} &
\includegraphics[height=\turnheightnew]{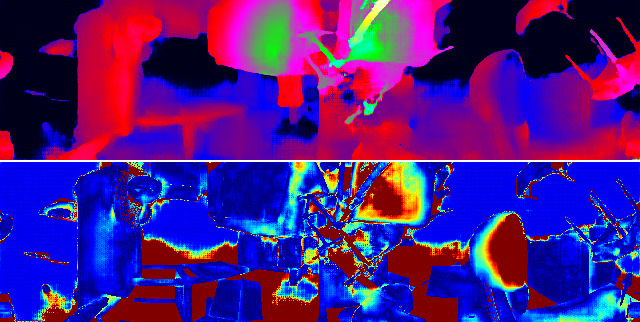} &
\includegraphics[height=\turnheightnew]{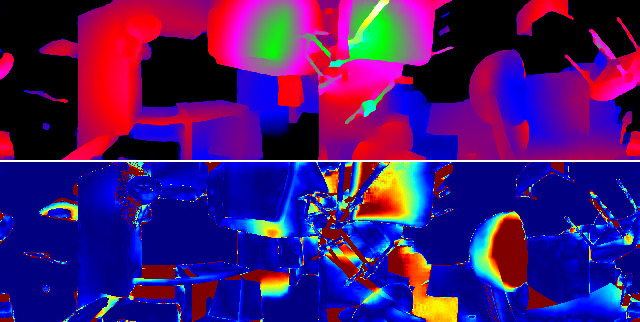} &
\includegraphics[height=\turnheightnew]{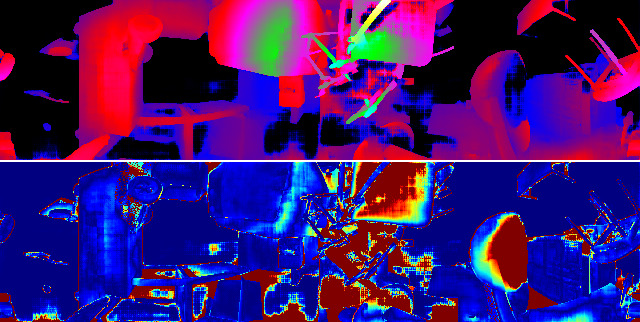} &
\includegraphics[height=\turnheightnew]{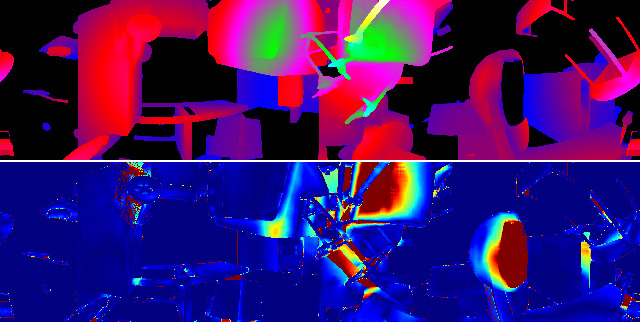}
\\

{\rotatebox{90}{\hspace{1.0mm}\scriptsize OmniThings}} &
\includegraphics[height=\turnheightnew]{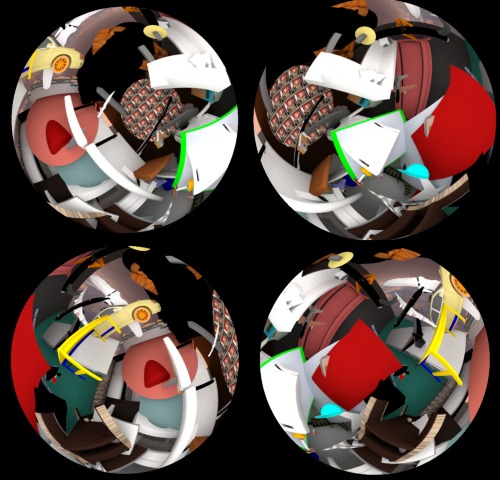} &
\includegraphics[height=\turnheightnew]{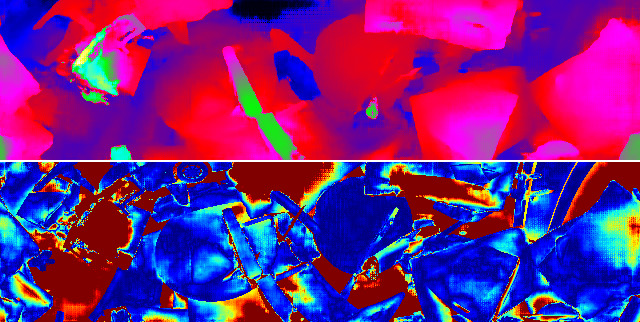} &
\includegraphics[height=\turnheightnew]{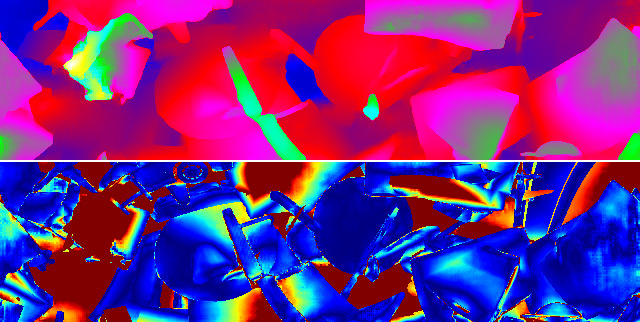} &
\includegraphics[height=\turnheightnew]{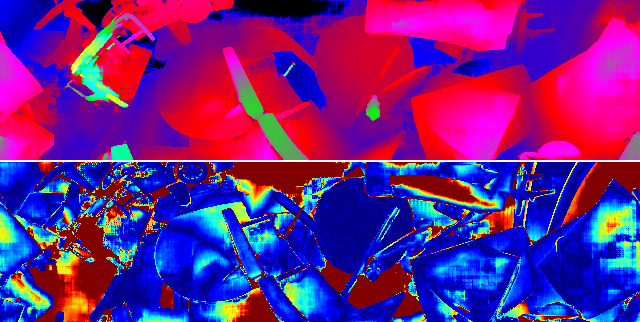} &
\includegraphics[height=\turnheightnew]{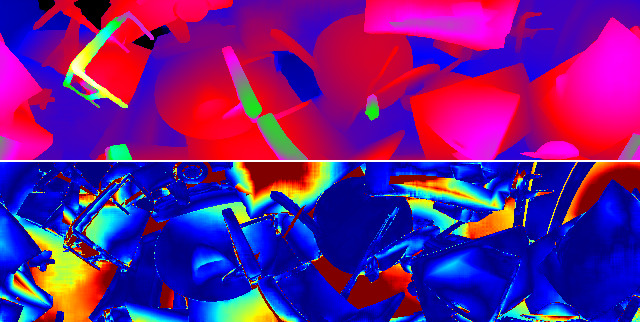}
\\
{\rotatebox{90}{\hspace{1.0mm}\scriptsize OmniThings}} &
\includegraphics[height=\turnheightnew]{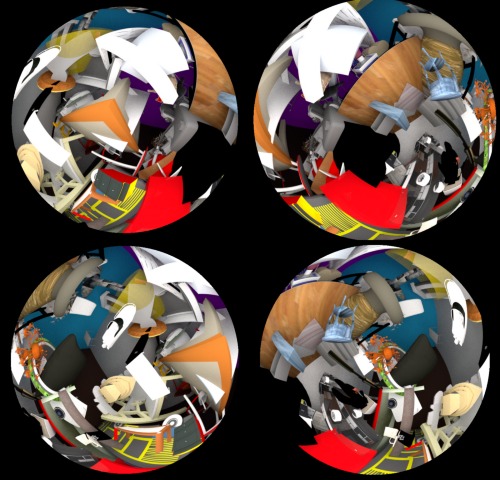} &
\includegraphics[height=\turnheightnew]{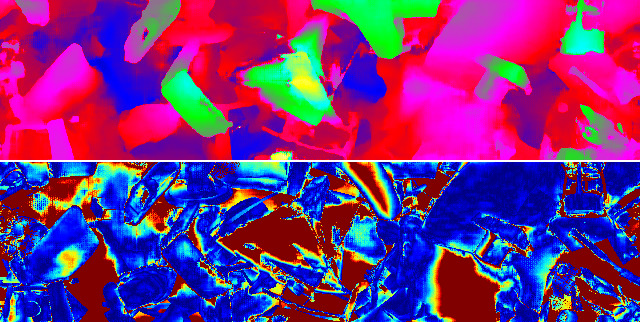} &
\includegraphics[height=\turnheightnew]{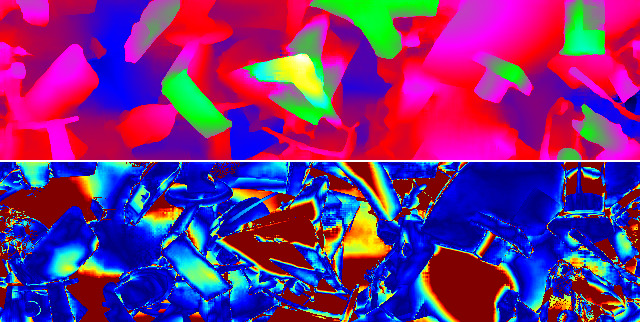} &
\includegraphics[height=\turnheightnew]{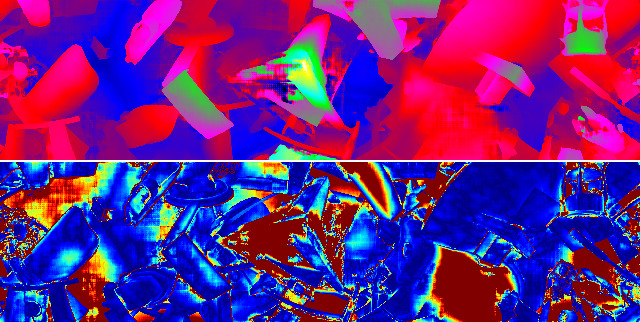} &
\includegraphics[height=\turnheightnew]{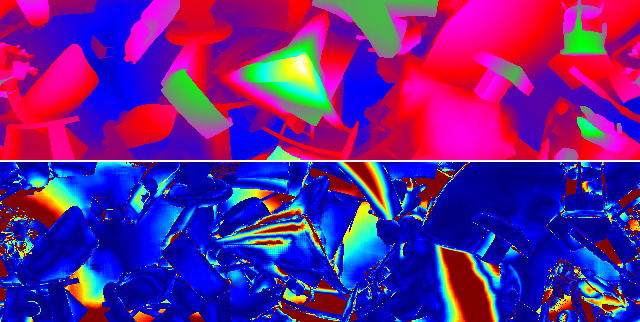}
\\

{\rotatebox{90}{\hspace{1.6mm}\scriptsize Omnihouse}} &
\includegraphics[height=\turnheightnew]{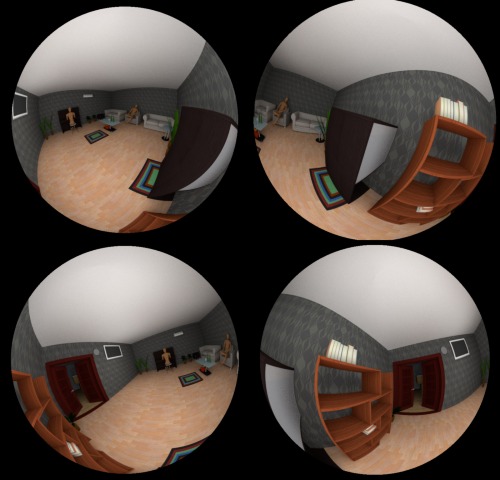} &
\includegraphics[height=\turnheightnew]{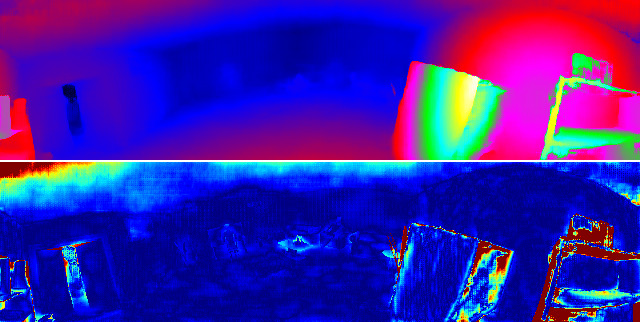} &
\includegraphics[height=\turnheightnew]{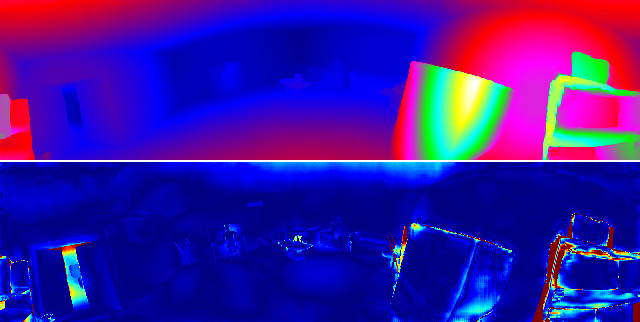} &
\includegraphics[height=\turnheightnew]{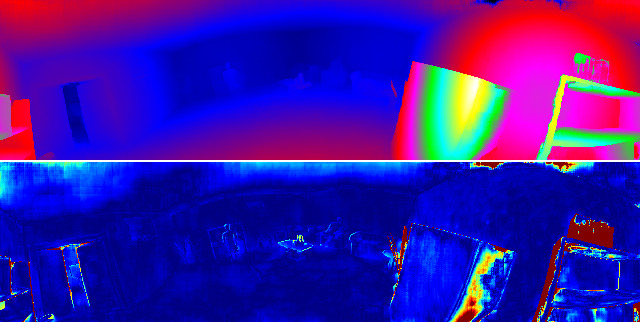} &
\includegraphics[height=\turnheightnew]{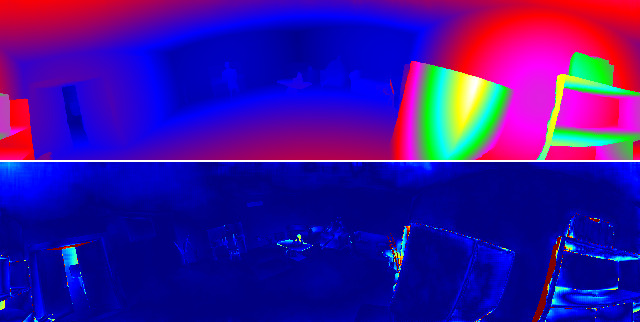}
\\

{\rotatebox{90}{\hspace{1.6mm}\scriptsize Omnihouse}} &
\includegraphics[height=\turnheightnew]{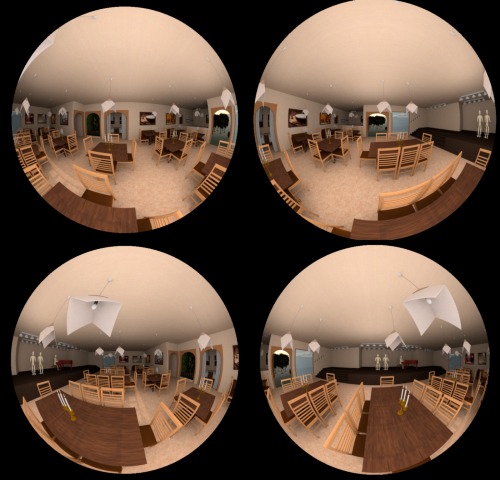} &
\includegraphics[height=\turnheightnew]{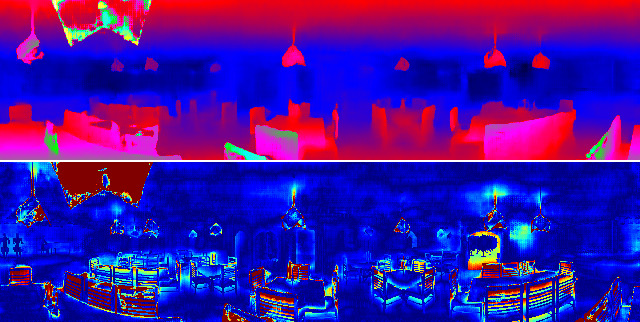} &
\includegraphics[height=\turnheightnew]{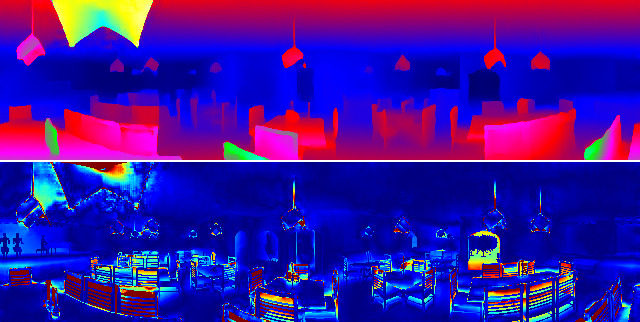} &
\includegraphics[height=\turnheightnew]{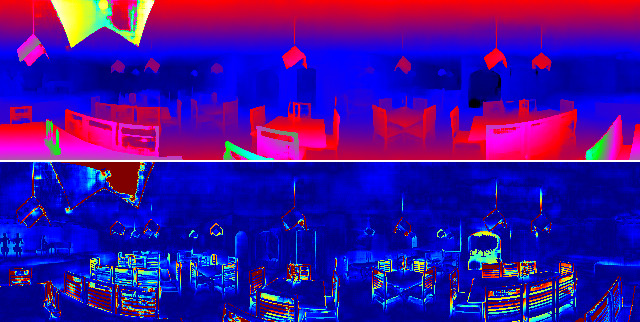} &
\includegraphics[height=\turnheightnew]{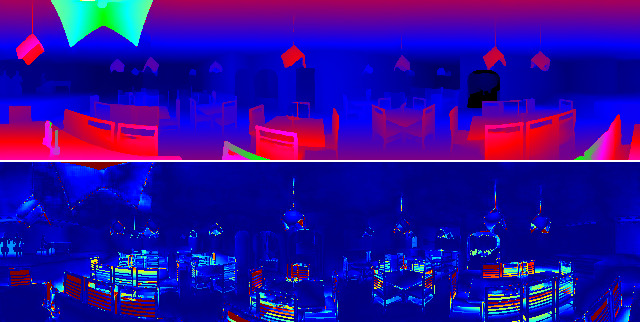}
\\
{\rotatebox{90}{\hspace{1.6mm}\scriptsize Omnihouse}} &
\includegraphics[height=\turnheightnew]{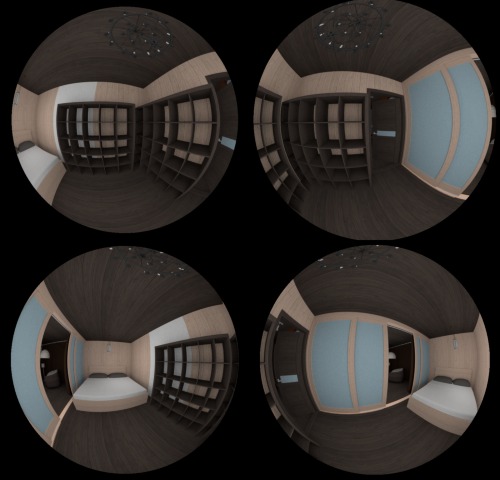} &
\includegraphics[height=\turnheightnew]{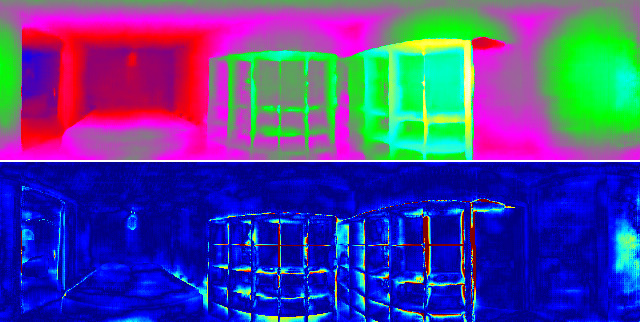} &
\includegraphics[height=\turnheightnew]{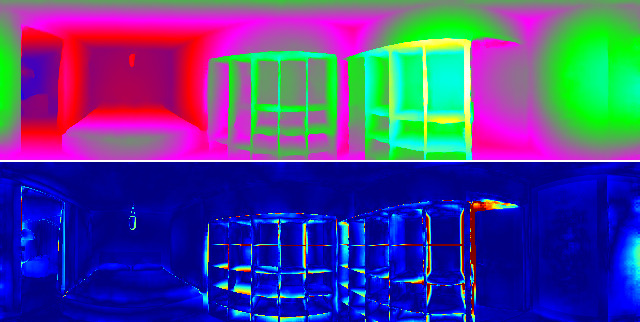} &
\includegraphics[height=\turnheightnew]{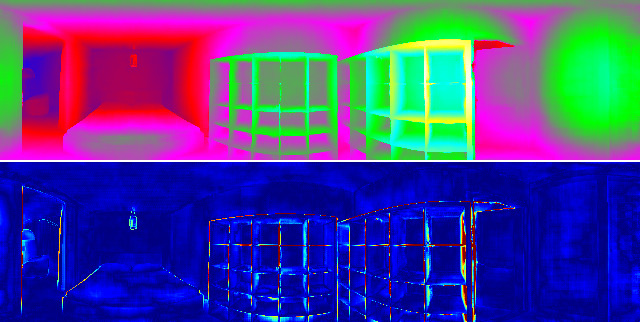} &
\includegraphics[height=\turnheightnew]{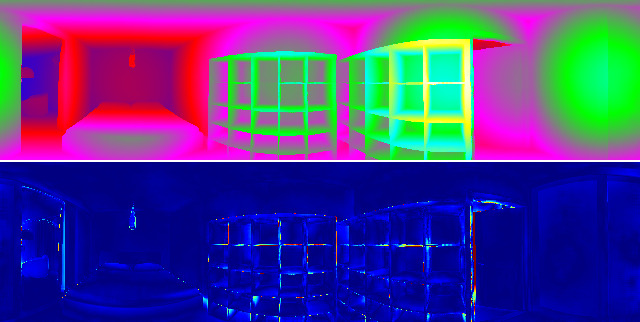}
\\

{\rotatebox{90}{\hspace{5.0mm}\scriptsize Suuny}} &
\includegraphics[height=\turnheightnew]{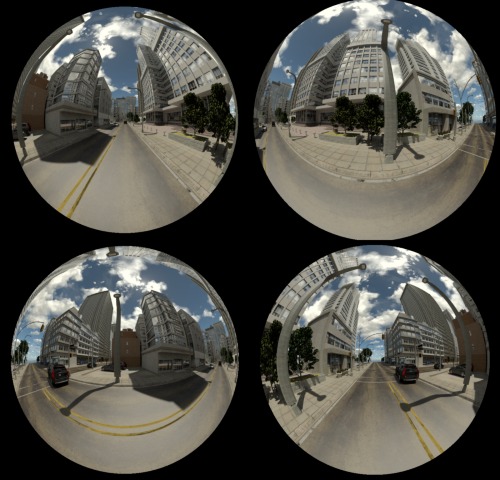} &
\includegraphics[height=\turnheightnew]{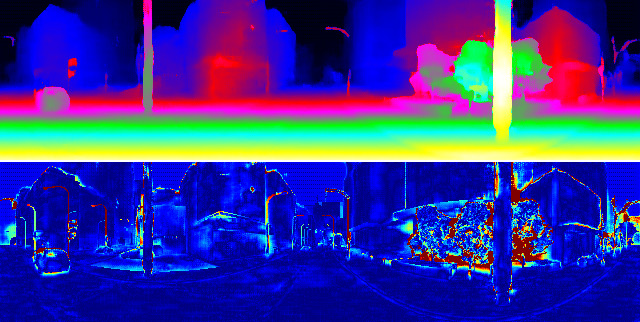} &
\includegraphics[height=\turnheightnew]{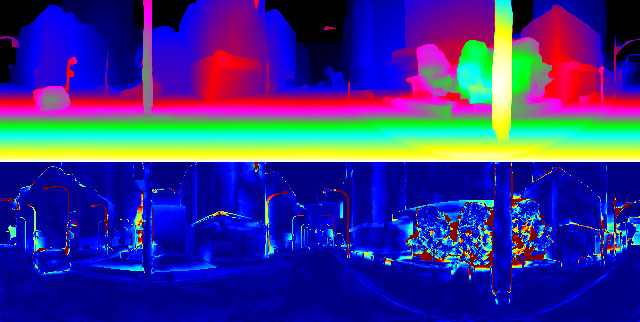} &
\includegraphics[height=\turnheightnew]{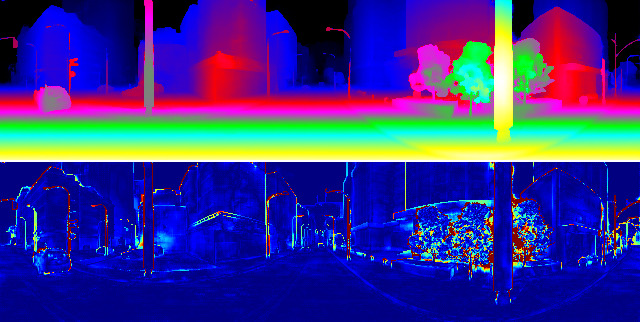} &
\includegraphics[height=\turnheightnew]{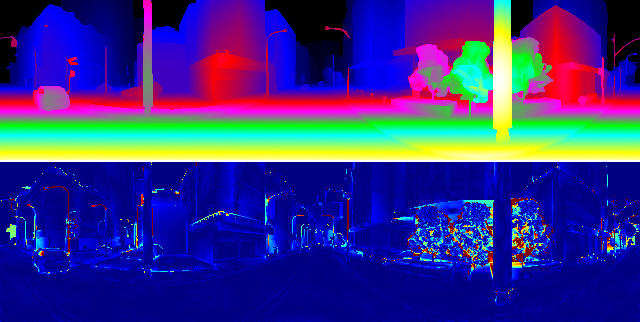}
\\
{\rotatebox{90}{\hspace{5.0mm}\scriptsize Suuny}} &
\includegraphics[height=\turnheightnew]{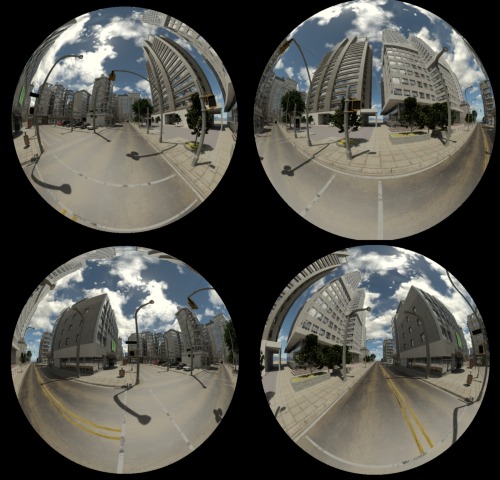} &
\includegraphics[height=\turnheightnew]{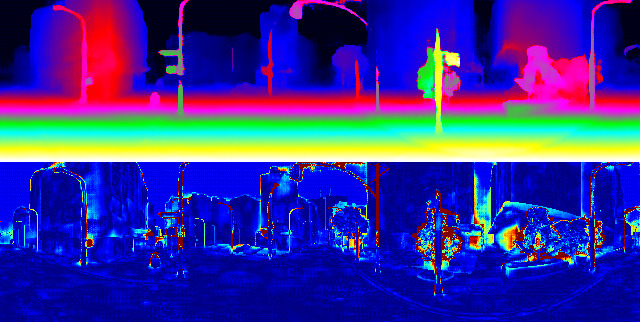} &
\includegraphics[height=\turnheightnew]{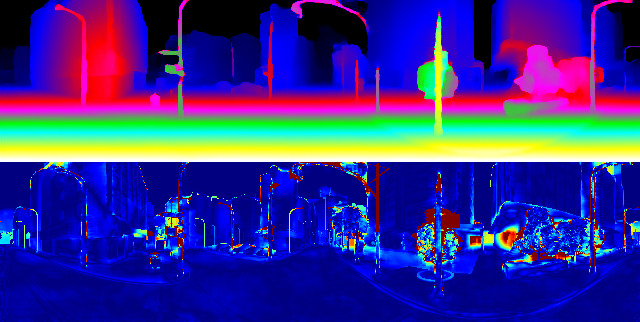} &
\includegraphics[height=\turnheightnew]{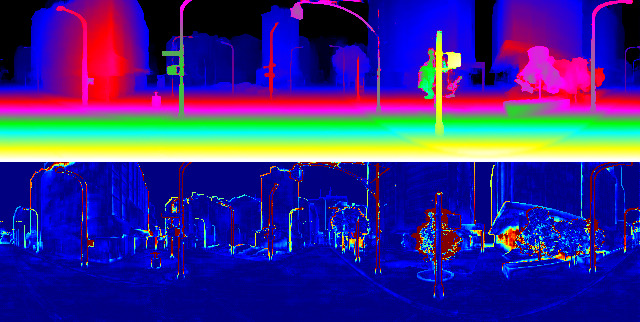} &
\includegraphics[height=\turnheightnew]{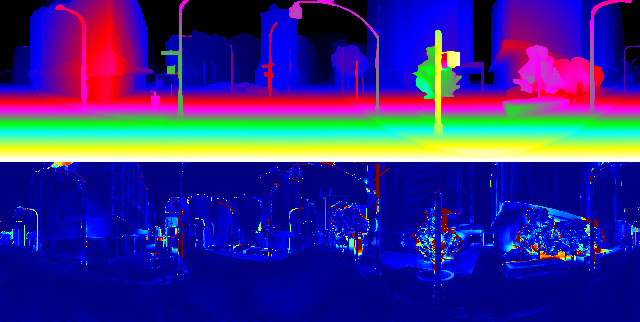}
\\
{\rotatebox{90}{\hspace{5.0mm}\scriptsize Suuny}} &
\includegraphics[height=\turnheightnew]{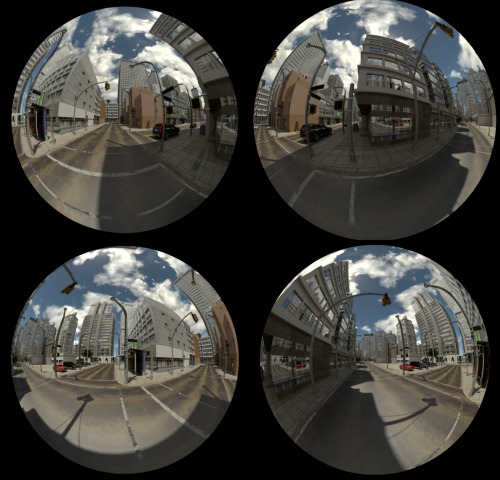} &
\includegraphics[height=\turnheightnew]{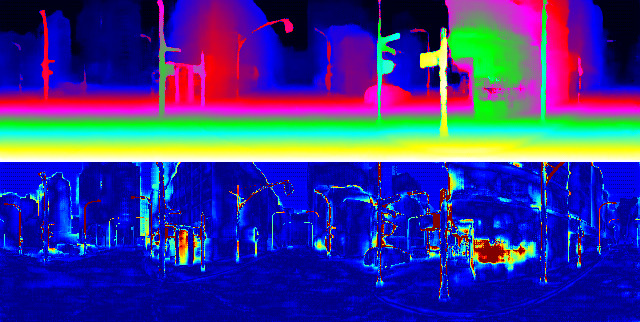} &
\includegraphics[height=\turnheightnew]{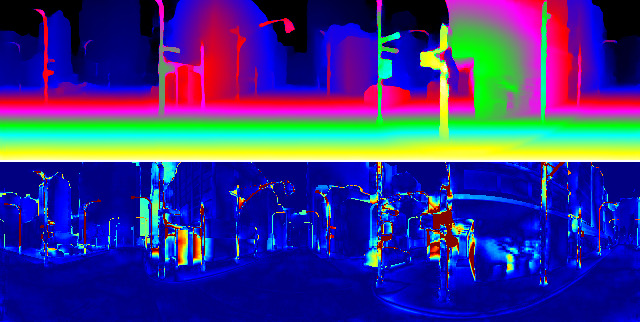} &
\includegraphics[height=\turnheightnew]{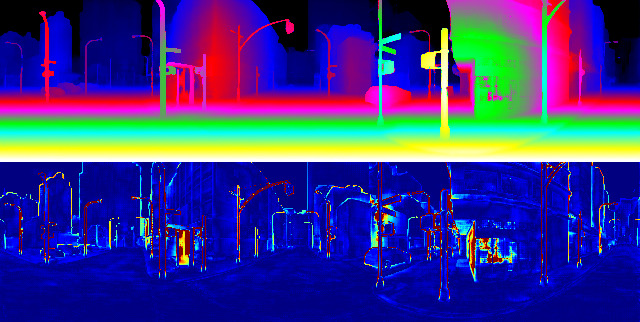} &
\includegraphics[height=\turnheightnew]{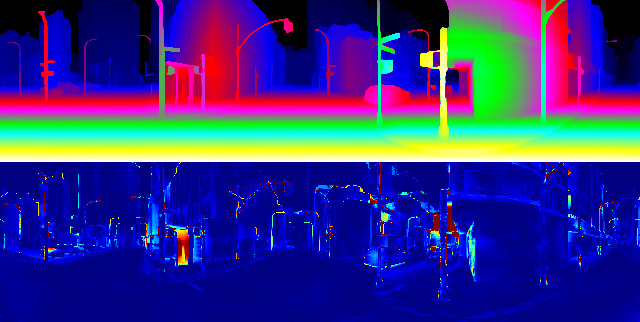}
\vspace{1pt}
\\
{\rotatebox{90}{\hspace{2.0mm}\scriptsize Real Scene}} &
\includegraphics[height=\turnheightnew]{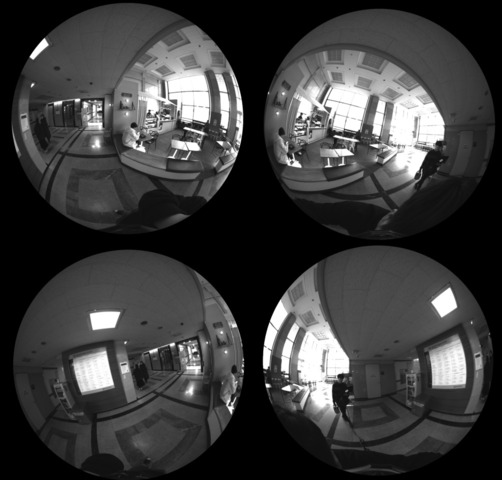} &
\includegraphics[height=\turnheightnew]{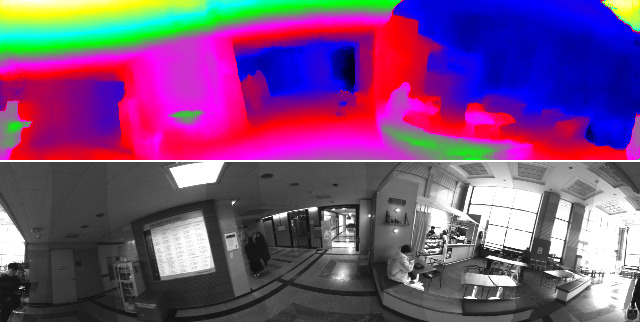} &
\includegraphics[height=\turnheightnew]{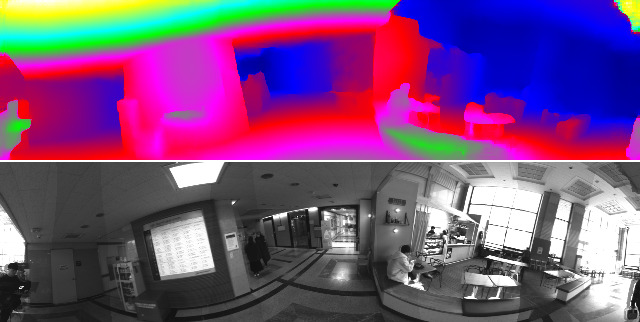} &
\includegraphics[height=\turnheightnew]{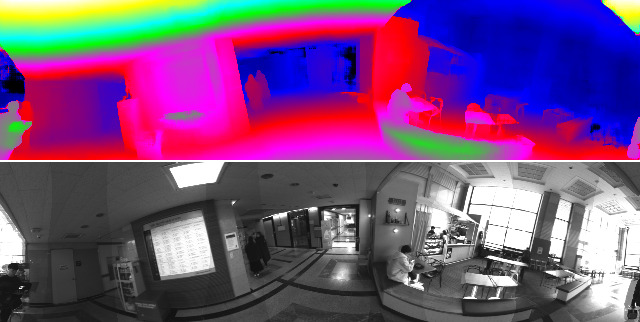} &
\includegraphics[height=\turnheightnew]{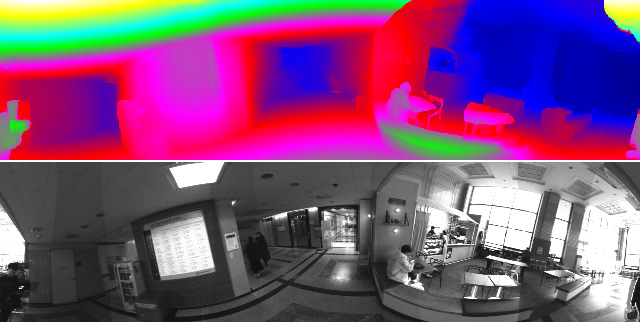}
\\
{\rotatebox{90}{\hspace{2.0mm}\scriptsize Real Scene}} &
\includegraphics[height=\turnheightnew]{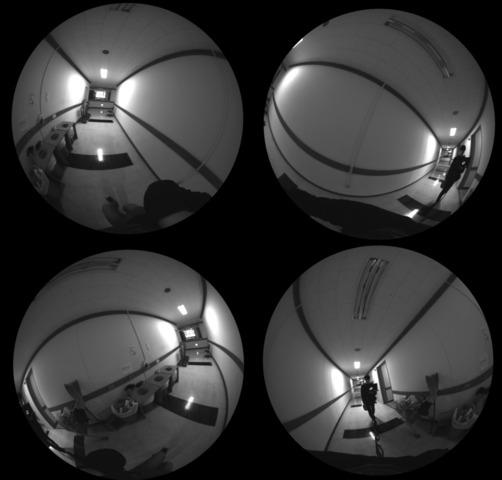} &
\includegraphics[height=\turnheightnew]{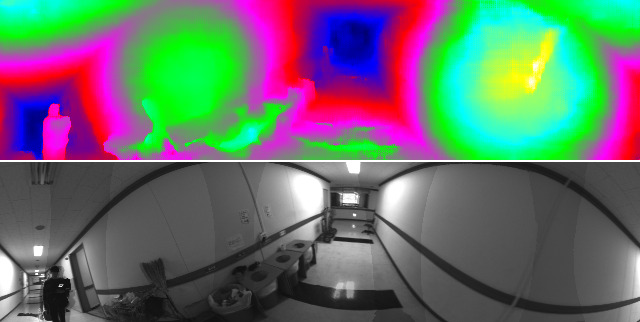} &
\includegraphics[height=\turnheightnew]{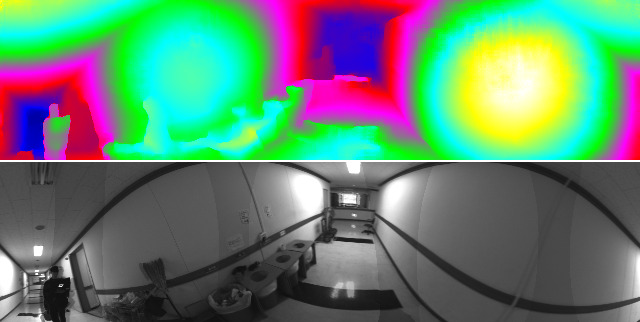} &
\includegraphics[height=\turnheightnew]{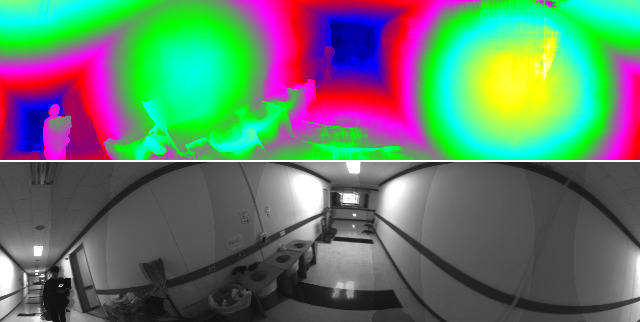} &
\includegraphics[height=\turnheightnew]{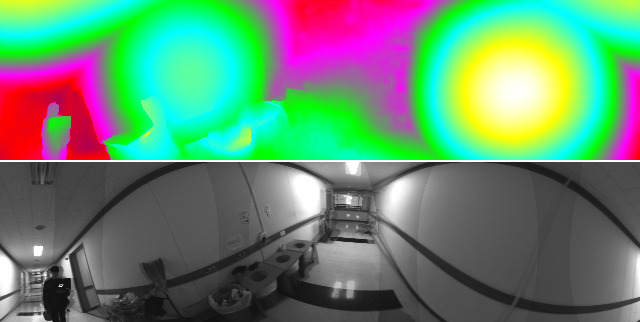}
\\
{\rotatebox{90}{\hspace{2.0mm}\scriptsize Real Scene}} &
\includegraphics[height=\turnheightnew]{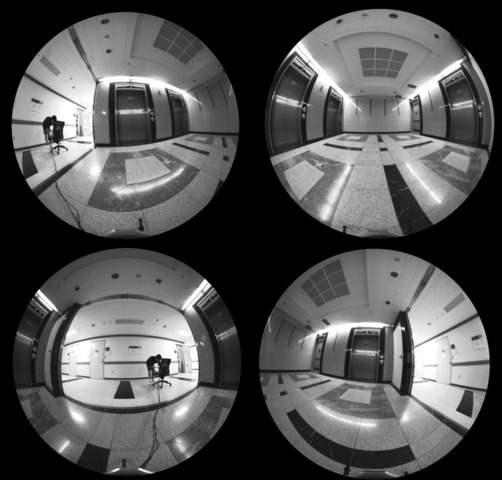} &
\includegraphics[height=\turnheightnew]{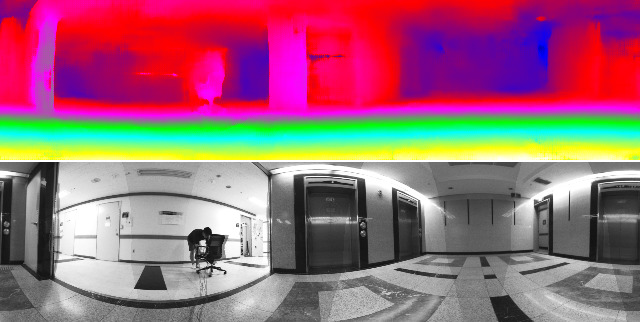} &
\includegraphics[height=\turnheightnew]{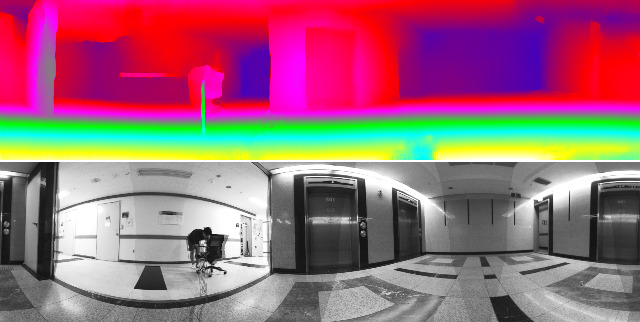} &
\includegraphics[height=\turnheightnew]{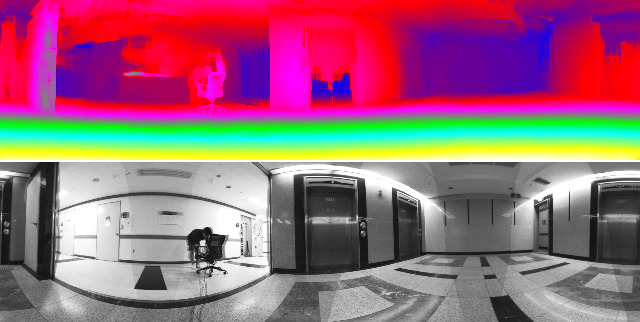} &
\includegraphics[height=\turnheightnew]{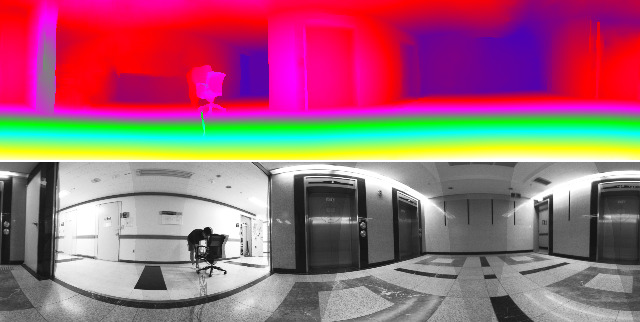}
\\

&
\scriptsize Input &
\scriptsize OmniMVS${}^+_4$-ft &
\scriptsize RomniStereo${}_4$-ft &
\scriptsize OmniMVS${}^+_{32}$-ft &
\scriptsize RomniStereo${}_{64}$-ft \\
\end{tabular}
 }
\vspace{0pt}
\caption{\small{Qualitative Comparison. Three examples from OmniThings, OmniHouse, Sunny, and real indoor data provided by OmniMVS are shown from top to bottom. Both the smallest and biggest versions of OmniMVS-ft and RomniStereo-ft are compared. The leftmost column is the input images. For synthetic samples, the results for each model include the estimated depth map and the error map. For the real samples, the results contain the predicted depth map and the resulting panorama. \footnotesize{\bf The images are best viewed in color and zooming in.}}}
\label{fig:comp}
\vspace{0pt}
\end{figure*}

\subsubsection{Main Comparison}
\label{sec:mc}

The quantitative comparison with prior arts is listed in Tab.~\ref{tab:quan}.  
{The results of OmniMVS~\cite{won2019omnimvs, won2020end} and S-OmniMVS~\cite{chen2023s} are mostly taken from their papers.}
As the journal version  OmniMVS${}^+$~\cite{won2020end} does not present the results of fine-tuning on OmniHouse and Sunny but its code repository\footnote{\url{https://github.com/hyu-cvlab/omnimvs-pytorch}} provides the model's parameters of \href{https://bit.ly/42h52fP}{OmniMVS${}^+_4$-ft} and \href{http://bit.ly/3TkLy64}{OmniMVS${}^+_{32}$-ft}, we run evaluation and include their results. The non-learning based Sphere-Stereo originally designed for short-baseline omnidirectional stereo matching provides code\footnote{\url{https://github.com/KAIST-VCLAB/sphere-stereo}}, so we test it on the datasets used here to reveal its effectiveness on the long-baseline condition. 

{The results indicate that Sphere-Stereo falls behind other learning-based methods, indicating the necessity of learning-based approaches for long-baseline omnidirectional stereo matching. } {When compared with the latest S-OmiNVS~\cite{chen2023s}, RomniStereo still shows apparent advantages in most cases. When the models are only trained on OmniThings, our best model RomniStereo${}_{64}$ performs better not only on in-domain test (17.6\% MEA improvement on OT), but also on cross-dataset generalization (30.2\% and 32.4\% MAE improvement on OH and Sn). When finetuned on Omnihouse and Sunny, although RomniStereo${}_{64}$  performs similarly to S-OmniNVS on OnmiHouse, it maintains the advantage on Sunny. }

{As the OmniMVS is the baseline of RomniStereo, the primary comparison lies in the different versions of OmniMVS$^+$~\cite{won2020end} and RomniStereo}.
The time complexity of RomniStereo is clearly smaller than OmniMVS$^+$ under the same $C$, and with the increase of $C$, RomniStereo's inference time grows much slower than OmniMVS$^+$. RomniStereo${}_{4}$ is just slightly faster to OmniMVS${}_{4}$ (0.09s vs 0.11s). However, RomniStereo${}_{8}$ is {twice as fast as} OmniMVS${}^+_{8}$ (0.10s vs 0.19s), and even faster than OmniMVS${}^+_{4}$. Finally, RomniStereo${}_{32}$ and RomniStereo${}_{64}$ just takes above quarter and half the time of OmniMVS${}^+_{32}$, respectively. 
Note that the similar runtimes of RomniStereo${}_{4}$ and RomniStereo${}_{8}$ can be attributed to the recurrent update, which consumes the majority of the runtime. This is because the dimension of the sampled correlation feature map is 36 (see Sec.~\ref{sec:reu}).

Under the same setting, RomniStereo outperforms OmniMVS$^+$ in most cases. In the smallest models without fine-tuning (RomniStereo${}_{4}$ vs. OmniMVS${^+}_{4}$), the MAE improvements on the 5 datasets, are 15.2\% (OT), 29.3\%(OH), 14.5\%(Sn), 17.9\%(Cd) and 14.8\%(Ss). Comparing the largest models (RomniStereo${}_{64}$ vs. OmniMVS${}^+_{32}$), the MAE improvements are still significant, being 10.9\% (OT), 43.1\%(OH), 14.8\%(Sn), 16.1\%(Cd) and 17.2\%(Ss).
When it comes to the smallest models with fine-tuning (RomniStereo${}_{4}$-ft vs. OmniMVS${}^+_{4}$-ft), the improvements become less but noticeable, which are 5.3\% (OT), 10.5\%(OH), 7.7\%(Sn), 1.6\%(Cd) and 9.1\%(Ss). However, a more fair comparison should be RomniStereo${}_{8}$-ft vs. OmniMVS${}^+_{4}$-ft, as they have a closer running time, and the MAE improvements are significant 22.0\% (OT), 31.4\%(OH), 26.2\%(Sn), 21.3\%(Cd) and 24.2\%(Ss).  Furthermore, when comparing the biggest fine-tuning models, \ie, RomniStereo${}_{64}$-ft vs. OmniMVS${}^+_{32}$-ft, the improvements are rather remarkable, being 46.6\% (OT), 34.4\%(OH), 43.9\%(Sn), 37.0\%(Cd) and 41.4\%(Ss), and the average reaches 40.7\%.

The qualitative comparison is shown in Fig.~\ref{fig:comp}. 
For OmniThings and OmniHouse, The predicted depth maps of RomniStereo-ft are more accurate and contain much fewer artifacts than those of OmniMVS$^+$-ft.  When it comes to Sunny, there seem to be no clear advantages of RomniStereo-ft over OmniMVS$^+$-ft on the output depth maps, as Sunny presents less complicated structures than the former two datasets.  However, from the error maps, one can observe that RomniStereo-ft produces fewer errors on the road area than OmniMVS-ft. For the real data samples, RomniStereo-ft still produces cleaner and more accurate depth maps, especially for the close-range region, which is crucial for robot navigation.  Admittedly, RomniStereo-ft can produce less accurate estimates on some far regions than OmniMVS-ft, for example, RomniStereo-ft tends to underestimate the distance of the end of the corridor of the first real example.

\section{CONCLUSIONS}

In this paper, we have presented an efficient and effective recurrent model for omnidirectional stereo matching by overcoming the difficulties in extending the advanced RAFT paradigm to the OSM domain. To close the gap between OSM and conventional pinhole image matching, we leveraged the camera structure prior to adaptively combining the opposite view to construct the reference/target volumes for later recurrent processing. Besides, we have also introduced two beneficial techniques, grid embedding and adaptive context feature generation to our RomniStereo model. Extensive experiments have demonstrated the effectiveness and efficiency of the proposed approach. In the future, we would like to investigate how to accelerate the model to a real-time level without sacrificing accuracy.

{\small
\bibliographystyle{IEEEtran}
\bibliography{ref}
}

\end{document}